\documentclass[lettersize,journal]{IEEEtran}
\usepackage{amsmath,amsfonts}
\usepackage{algorithmic}
\usepackage{array}
\usepackage[caption=false,font=normalsize,labelfont=sf,textfont=sf]{subfig}
\usepackage{textcomp}
\usepackage{stfloats}
\usepackage{url}
\usepackage{verbatim}
\usepackage{graphicx}
\usepackage{bigstrut,multirow}
\usepackage{amsmath}
\usepackage{amsfonts}
\usepackage{booktabs}
\usepackage{orcidlink} 
\def\BibTeX{{\rm B\kern-.05em{\sc i\kern-.025em b}\kern-.08em
    T\kern-.1667em\lower.7ex\hbox{E}\kern-.125emX}}
\usepackage{balance}
\begin{document}
\title{PVLM: Parsing-Aware Vision-Language Model with Dynamic Contrastive Learning for Zero-Shot Deepfake Attribution}
\author{Yaning Zhang \orcidlink{0000-0001-8442-2777}, Jiahe Zhang, Chunjie Ma \orcidlink{0000-0002-6348-671X}, Weili Guan \orcidlink{0000-0002-5658-5509}, \emph{Member, IEEE}, Tian Gan \orcidlink{0000-0003-3197-5698}, Zan Gao \orcidlink{0000-0003-2182-5741}, \emph{Senior Member, IEEE}
		\thanks{This work was supported in part by the National Natural Science Foundation of China (No.U25A20444, No.62372325, No.62402255), Natural Science Foundation of Tianjin Municipality (No.23JCZDJC00280), Shandong Provincial Natural Science Foundation (No.ZR2024QF020), Shandong Province National Talents Supporting Program (No.2023GJJLJRC-070), Shandong project towards the integration of education and industry (No.801822020100000024), Young Talent of Lifting engineering for Science and Technology in Shandong (No. SDAST2024QTB001), Shandong Project towards the Integration of Education and Industry (No.2024ZDZX11) (Corresponding author: Zan Gao)} 
\thanks{Y. Zhang is with Faculty of Computer Science and Technology, Qilu University of Technology (Shandong Academy of Sciences), Jinan, 250014, China. E-mail: zhangyaning0321@163.com}
\thanks{J. Zhang is with Shandong University of Science and Technology, Jinan, 250031, China. E-mail: zhangjiahe00600@163.com.}
 \thanks{C. Ma is with the Shandong Artificial Intelligence Institute, Qilu University of
	Technology (Shandong Academy of Sciences), Jinan, 250014, China. E-mail: mcj@machunjie.com }
\thanks{W. Guan is with the School of Electronics and Information Engineering, Harbin Institute of Technology, Shenzhen 150001, China. E-mail: guanweili@hit.edu.cn}

\thanks{T. Gan is with the School of Computer Science and Technology, Shandong University, Qingdao, 266237, China. E-mail: gantian@sdu.edu.cn}

\thanks{Z. Gao is with Shandong Artificial Intelligence Institute, Qilu University of Technology (Shandong Academy of Sciences), Jinan, 250014, China, and also with the Key Laboratory of Computer Vision and System, Ministry of Education, Tianjin University of Technology, Tianjin, 300384, China. E-mail: zangaonsh4522@gmail.com } 
}
\maketitle

\begin{abstract}
	The challenge of tracing the source attribution of forged faces has gained significant attention due to the rapid advancement of generative models. However, existing deepfake attribution (DFA) works primarily focus on the interaction among various domains in vision modality, and other modalities such as texts and face parsing are not fully explored. Besides, they tend to fail to assess the generalization performance of deepfake attributors to unseen advanced generators like diffusion in a fine-grained manner. In this paper, we propose a novel \underline{\textbf{p}}arsing-aware \underline{\textbf{v}}ision \underline{\textbf{l}}anguage \underline{\textbf{m}}odel with dynamic contrastive learning (PVLM) method for \underline{\textbf{z}}ero-\underline{\textbf{s}}hot \underline{\textbf{d}}eep\underline{\textbf{f}}ake \underline{\textbf{a}}ttribution (ZS-DFA), which facilitates effective and fine-grained traceability to unseen advanced generators. Specifically, we conduct a novel and fine-grained ZS-DFA benchmark to evaluate the attribution performance of deepfake attributors to unseen advanced generators like diffusion. Besides, we propose an innovative PVLM attributor based on the vision-language model to capture general and diverse attribution features. We are motivated by the observation that the preservation of source face attributes in facial images generated by GAN and diffusion models varies significantly. We propose to employ the inherent facial attributes preservation differences to capture face parsing-aware forgery representations. Therefore, we devise a novel parsing encoder to focus on global face attribute embeddings, enabling parsing-guided DFA representation learning via dynamic vision-parsing matching. Additionally, we present a novel deepfake attribution contrastive center loss to pull relevant generators closer and push irrelevant ones away, which can be introduced into DFA models to enhance traceability. Experimental results show that our model exceeds the state-of-the-art on the ZS-DFA benchmark via various protocol evaluations. Codes will be available at GitHub.
\end{abstract}


\section{Introduction}
\vspace{-0.3em}

\begin{figure}[t!]
	\centering
	\includegraphics[width=\linewidth]{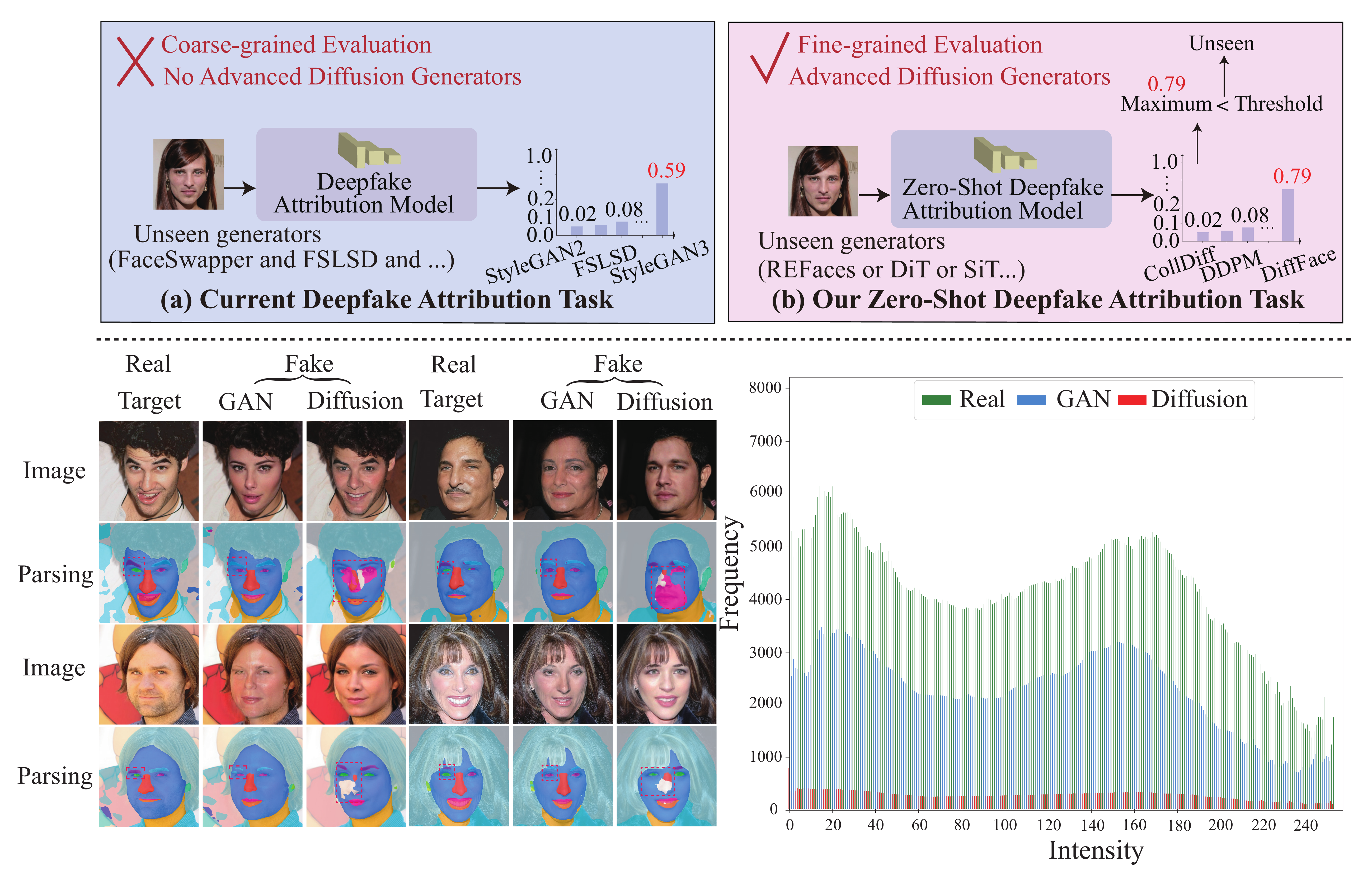} 
	\vspace{-2em}
	\caption{ Up: The overview of the deepfake attribution task. Down: Illustration of the face attribute-sensitive trait. Left: The visualization of real, GAN, and diffusion face parsing images. Right: The feature distribution histogram of real, GAN, and diffusion face parsing images. }\label{fig1}
\end{figure}

With the proliferation of advanced generative frameworks, including generative adversarial networks (GANs) \cite{GAN} and diffusion models \cite{diff}, maliciously fabricated deepfake content on social platforms has raised significant concerns over facial integrity and personal privacy \cite{ deepfakesurvey}. As a result, tracing the source of deepfakes has become a critical and urgent issue. Deepfake attribution (DFA) aims to identify and attribute forged faces using deep learning-based methods. Although current DFA techniques \cite{DEFAKE, DNADet} have made significant progress, they predominantly assume a closed-world setting where the training and testing sets share identical category distributions. This assumption limits their applicability in open-world scenarios, where novel forgery attacks continuously emerge. To overcome this problem, open-world deepfake attribution (OW-DFA) leverages unlabeled data in open-world contexts, to enhance attribution performance for forgery faces created by both seen and unseen generators. However, most DFA models \cite{CPL, MPSL} focus solely on the interaction between image and frequency information within the visual modal, neglecting other modalities such as fine-grained text and face parsing, which may hinder generalization to unseen generators. For instance, DNA-Det \cite{DNADet} identifies globally consistent forgery features using the image modality for GAN architecture attribution. MPSL \cite{MPSL} further explores multi-scale image and frequency relationships among samples, enabling the attribution of both GAN and diffusion-generated face images \cite{DFFD}. However, they struggle to evaluate the generalization to unseen generators in a fine-grained manner.

\begin{figure}[t!]
	\centering
	\includegraphics[width=\linewidth]{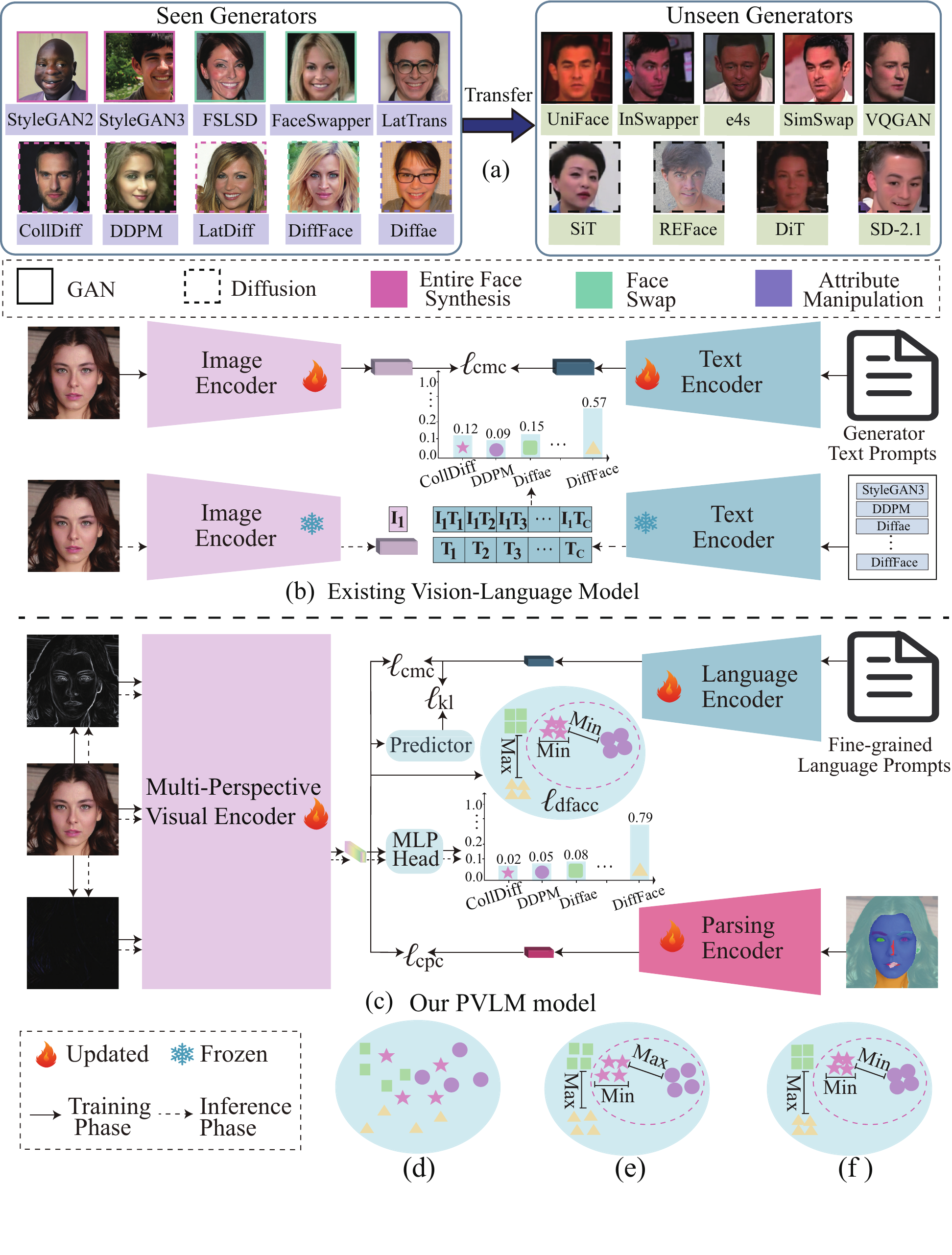} 
	\vspace{-4em}
	\caption{(a) In ZS-DFA tasks, training and testing datasets are derived from distinct domains, with generators in the testing dataset being unseen during training. (b) Existing vision-language foundation model CLIP. (c) The proposed PVLM model. (d) Original DFA feature space. (e) DFA feature space supervised by vanilla contrastive center loss. (f) DFA feature space supervised by our DFACC loss.}\label{fig2}
	\vspace{-2em}
\end{figure}

To alleviate the limitations, we introduce a novel task, namely, zero-shot deepfake attribution (ZS-DFA). As Fig.~\ref{fig1} shows, unlike current deepfake attribution tasks (See  Fig.~\ref{fig1} (a)) that conduct coarse-grained attribution performance evaluation on unseen GAN generators, our ZS-DFA task (See  Fig.~\ref{fig1} (b)) aims to identify advanced unseen generators such as diffusion, fine-grainedly. Specifically, as shown in Fig.~\ref{fig2} (a), ZS-DFA encompasses about 20 advanced generators, categorized into three manipulations such as entire face synthesis (EFS), face swap (FS), and attribute manipulation (AM). The challenge of ZS-DFA is how to design an attributor to mine general and discriminative forgery attribution traces synthesized by seen generators, to enhance the attribution performance for fake faces created by unseen advanced ones.

\begin{figure}[t!]
	\centering
	\includegraphics[width=\linewidth]{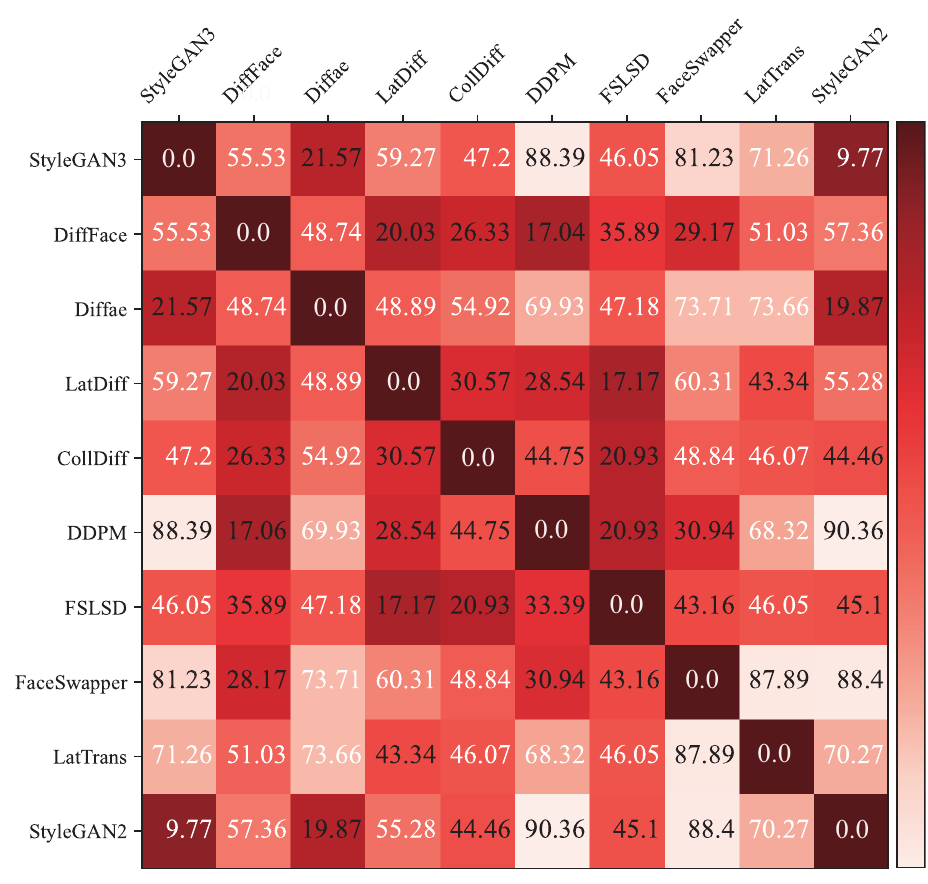} 
	\vspace{-2em}
	\caption{Cross-generator correlation matrix visualization. We randomly select 10k samples from each of the two generators to calculate the FID score, to measure the relevance. The lower the FID score, the greater the correlation across generators. The darker red means stronger correlation and the lighter red denotes weaker relevance.   }\label{figcm}
\end{figure}

\begin{figure*}[t]
	\centering
	\includegraphics[width=\linewidth]{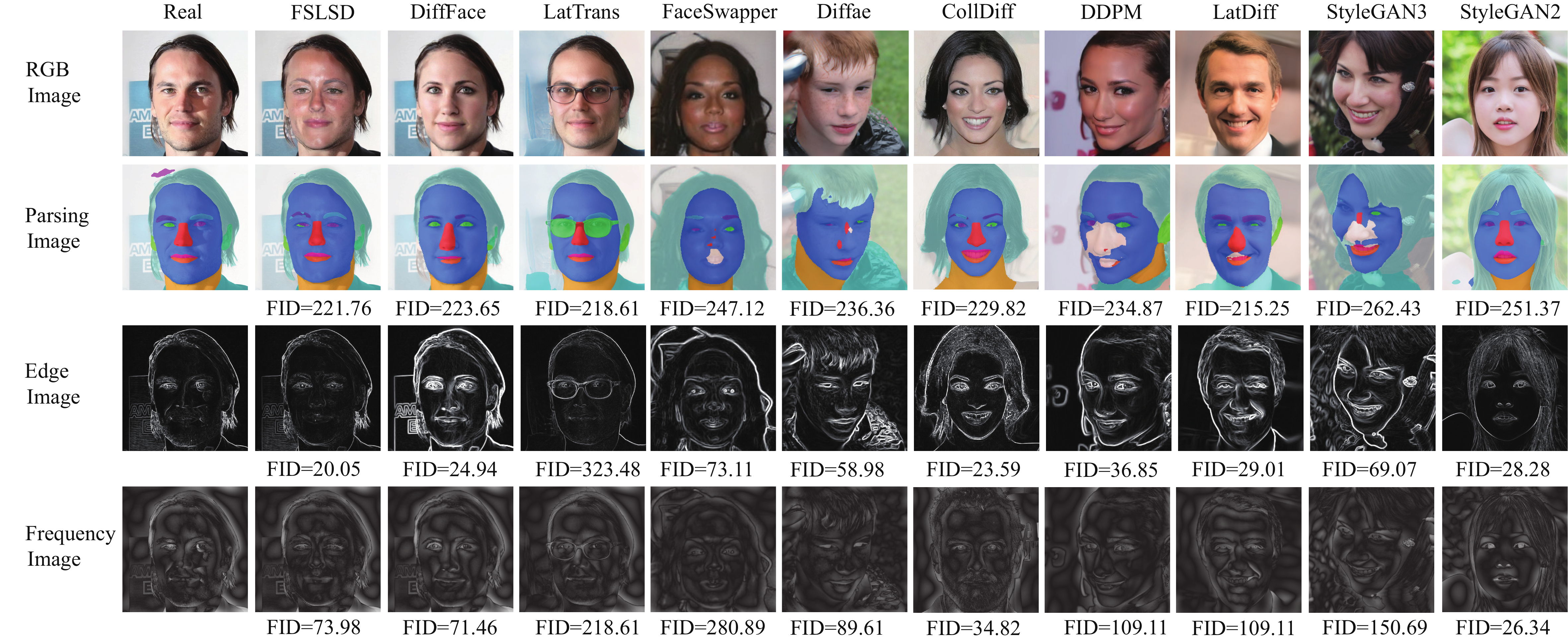} 
	\caption{The visualization of priors from different domains including face parsing, edge, and frequency. Each column shows a face yielded by various
			generators. The first to forth rows represent the RGB image, face parsing image, edge image extracted by Sobel, and the frequency image derived from the fast fourier transform (FFT), respectively. We randomly select 10k real prior images and 10k fake ones to calculate the FID score. The higher the FID score, the greater the difference between the real and fake prior distribution.}\label{figintro2}	
\end{figure*}

Inspired by the contrastive language-image pre-training (CLIP) model \cite{clip} which excels at zero-shot learning, we propose a parsing-aware vision-language model with dynamic contrastive learning (PVLM) framework. Unlike CLIP which only focuses on image and text modals for zero-shot classification in general natural image scenarios, our PVLM model introduces fine-grained text and face parsing modalities, to enhance the learning of multiple-perspective (i.e., image, noise, and edge) visual features via cross-modality matching and dynamic cross-perspective alignment, to achieve fine-grained ZS-DFA. Specifically, our PVLM method differs from CLIP in the following points (See Fig.~\ref{fig2}): First, we design a multi-perspective visual encoder (MPVE) to explore global and diverse multi-view (i.e., image, edge, and noise) forgery representations via the multi-head interaction module (MMI), as there are significant discrepancies between real and fake images in both edge and noise views \cite{genface,MFCLIP}. Second, since most face manipulations such as FS struggle to preserve facial attributes of the source image like eyebrows and eyes \cite{deepfakesurvey}, as shown in Fig.~\ref{fig1}, we employ a face parser \cite{faceparser} to generate parsing images to focus on different facial attribute regions. We observe that there are significant differences between pristine and forged face parsing images (See Fig.~\ref{fig1}). Specifically, authentic face parsing images exhibit more precise attribute areas (e.g., eyes and eyebrows) than forged ones. Besides, parsing images derived from GAN-generated faces retain more accurate semantic attributes than those extracted from diffusion faces. In Fig.~\ref{figintro2}, the parsing image in the diffusion-generated facial image is more discriminative than that in the GAN-synthesized one, compared to other priors extracted by edge Sobel or the frequency-based FFT. The FID score of the diffusion-generated face parsing mode is significantly higher than that of other diffusion priors, and there are evident FID score differences in face parsing images among various generators, compared to other edge and frequency distributions. This demonstrates the generalization advantage of face parsing priors over other forgery cues (e.g., frequency-domain artifacts). Based on the above observation and considering that different generators may produce various facial attribute manipulation artifacts, we propose a parsing encoder (PE) to focus on global fine-grained and diverse semantic forgery regions of facial attributes, which can adapt the face parsing information to our DFA domain and align the face parsing forged features with multi-view visual ones, guiding our PVLM model to learn discriminative and general DFA embeddings. We present a novel dynamic cross-perspective contrastive loss to achieve dynamic vision-parsing matching. We also propose a language encoder to extract fine-grained language attribution embeddings for comprehensive vision-language matching. Finally, inspired by the vanilla contrastive center loss (VCC) \cite{contrastive}, we aim to enhance DFA representations through metric learning. However, VCC tends to enforce intra-generator compactness and inter-generator separability (see Fig.~\ref{fig2} (e)), which is regarded as a sub-optimal metric learning solution since there are strong correlations between some generators, with some exhibiting weak correlations (see Fig.~\ref{figcm}), and relevant samples across generators should be closer to each other in the feature space. To address the limitation, we design a novel deepfake attribution contrastive center (DFACC) loss (See Fig.~\ref{fig2} (f)) to minimize the distance between samples within the generator, while bringing related samples across generators closer and pushing unrelated ones farther apart, which facilitates our model to capture generator-common artifacts to improve the generalization. In summary, the contributions of our work are as follows:

$\bullet$ To the best of our knowledge, we conduct the first zero-shot fine-grained deepfake attribution benchmark, which contains novel and advanced generators like diffusion, to facilitate the ZS-DFA application in open-world scenarios where innovative face forgery generators emerge.

$\bullet$ We propose a novel parsing-aware vision-language model with dynamic contrastive learning (PVLM) deepfake attributor, which introduces fine-grained text and face parsing modalities to enhance the learning of visual forgery features across image, noise, and edge views, thus mining both general and discriminative attribution patterns.   

$\bullet$ We propose a plug-and-play deepfake attribution contrastive center criterion to realize flexible intra-generator compactness and inter-generator separability, which could be integrated into any deepfake attributors to improve traceability.

$\bullet$ Experiments conducted on our benchmark demonstrate that our PVLM method surpasses the state of the art in zero-shot deepfake attribution scenarios.

\section{Related Work}

{\bfseries\setlength\parindent{0em} Deepfake Attribution.} DFA aims to identify the source of deepfake images through deep learning-based models, which is a challenging task due to the sophistication of deepfake technology. Most existing efforts \cite{DNADet,DEFAKE,CPL,dfagan} focus on attributing GAN-generated content, with little attention given to more recent advanced generators like diffusion. DNA-Det \cite{DNADet} is designed to capture globally consistent architecture traces via pre-training on image transformation classification and patch-based contrastive learning. Sun et al. \cite{CPL} propose a contrastive pseudo learning (CPL) network for OW-DFA tasks, which evaluates attribution performance against various GAN-generated face images. To address the limitation, MPSL \cite{MPSL} models global multi-scale image and frequency relations among samples, to attribute fake images generated by diffusion models. DE-FAKE \cite{DEFAKE} employs the CLIP model to attribute fake images created by diffusion text-to-image generation models. However, they struggle to fully explore fine-grained text and vision modalities, and assess the generalization performance of deepfake attributors to unseen generators in a fine-grained manner. By contrast, we conduct the bi-modal (i.e., parsing and text) guided multi-view (i.e., image, edge, and noise) visual forgery representation learning, to realize fine-grained ZS-DFA.

{\bfseries\setlength\parindent{0em} Zero-Shot Learning.} Zero-Shot learning (ZSL) enables a model to recognize and classify instances from unseen classes during the training phase. Traditionally, ZSL \cite{Diffzs,CARZero} for image classification relies on learning auxiliary attributes rather than predefined class labels. Predictions are made by evaluating the similarity between image and text embeddings, derived from encoding both images and textual class descriptions \cite{FLIP}. Existing approaches like CLIP \cite{clip} tend to train image and text encoders to learn joint representations. Recent works \cite{clipzs,clipvlm} have focused on improving zero-shot classification performance of CLIP by enhancing textual descriptions. In this work, we are the first to attain excellent zero-shot performance for DFA. Unlike CLIP which focuses solely on image-text matching, our PVLM method captures diverse global visual forgery features across image, noise, and edge perspectives, performing both vision-parsing matching and vision-language alignment to explore comprehensive and model-agnostic forensic representations.

{\bfseries\setlength\parindent{0em} Vision-Language Model.} C2P-CLIP \cite{C2P-CLIP} presents a novel approach known as category common prompt CLIP, which injects category-specific prompts into the text encoder and enriches the image encoder with relevant category concepts via vision-language contrastive learning, resulting in enhanced detection performance. MFCLIP \cite{MFCLIP} integrates fine-grained noise forgery embeddings with global image forgery prompts, enhancing them via dynamic vision-language contrastive learning to boost generalization to face images generated by diffusion models. ForensicAdapter \cite{forensics} designs an adapter to capture unique blending boundaries of forged faces, and then enhances CLIP visual tokens using a specialized interaction strategy to boost knowledge transfer between CLIP and the adapter. By contrast, based on the novel finding of face parsing feature differences between GAN and diffusion models, we employ distinct parsing features and fine-grained text prompts to enhance the visual forgery representation learning across noise, image, and edge views, to realize fine-grained ZS-DFA.

\section{Methodology}
\begin{figure*}[t!]
\centering
	\includegraphics[width=\linewidth]{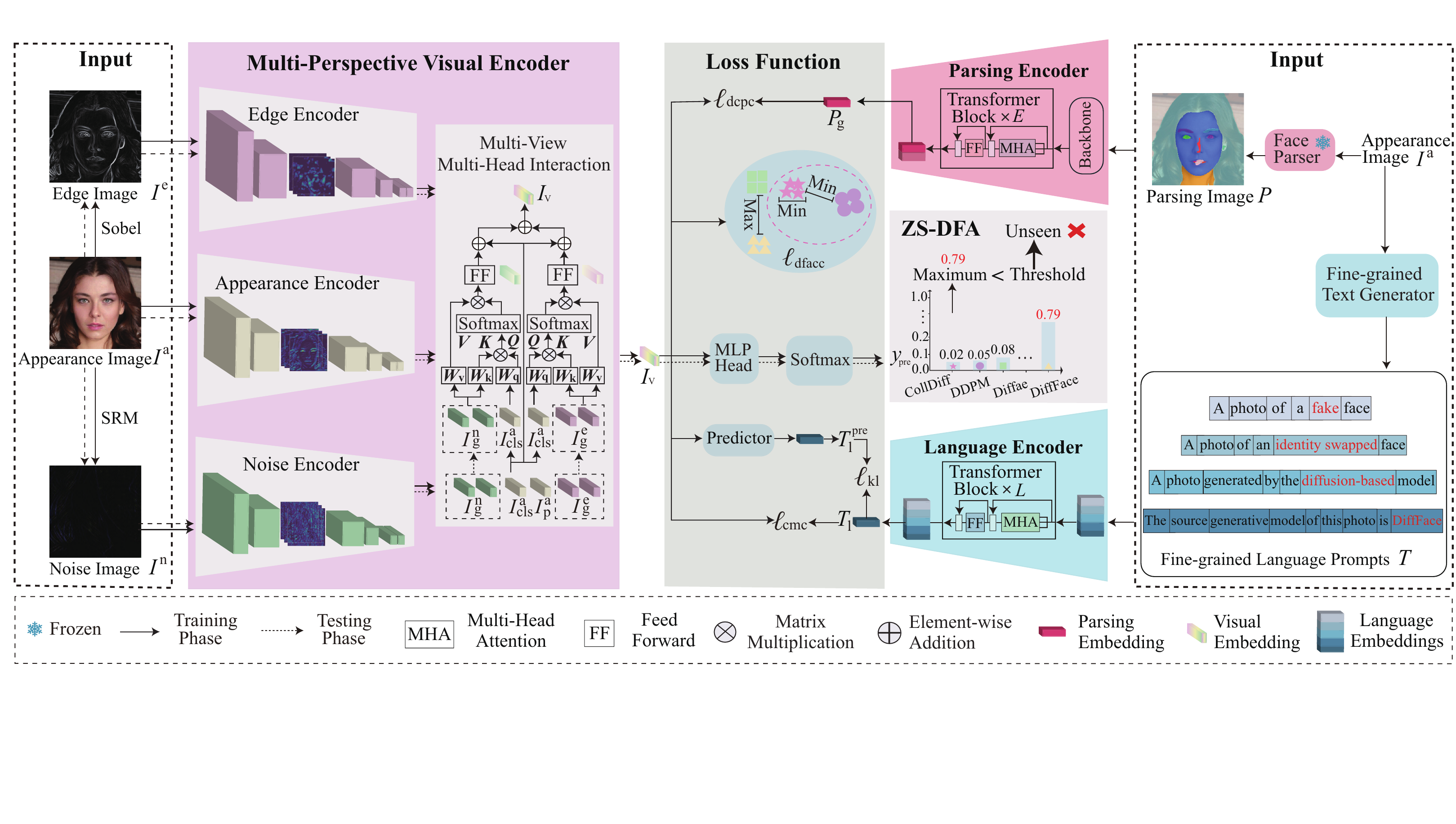} 
	\vspace{-7em}
	\caption{The workflow of our PVLM model to conduct ZS-DFA. We first send the appearance image to the Sobel, SRM operator, face parser, and fine-grained text generator, to derive the edge image, noise image, parsing image, and text prompts, respectively. We then fed the appearance, edge, and noise image into MPVE to extract visual deepfake attribution features across multiple views. Meanwhile, the face parsing image and text prompts are transferred to the PE and LE to acquire face attribute features and language embeddings, accordingly. We then conduct the vision-language matching, dynamic vision-parsing alignment, and flexible metric learning. Finally, multi-view visual features are imparted to the MLP head and softmax to yield the prediction.}
	\label{fig3}
\end{figure*}

\subsection{Problem Definition}
For ZS-DFA, given a seen domain $(X_\text{s},Y_\text{s})$ and an unseen domain $(X_\text{u},Y_\text{u})$, the input distributions of the seen and unseen domains are distinct, and there is no intersection between their label spaces, i.e. $X_\text{s} \neq X_\text{u}$ and $Y_\text{s}\cap Y_\text{u} = \emptyset$, where $X$ denotes the input distribution and $Y$ represents the label space. ZS-DFA aims to learn a deepfake attributor, trained on the seen domain, to classify the distribution from the unseen domain. In our work, we employ the training set $(I_\text{s}$, $y_\text{s}) \in \mathcal{D}_\text{tra}$ from the seen domain to train our model, and test it using the testing set $(I_\text{u}$, $y_\text{u}) \in \mathcal{D}_\text{tes}$ from the unseen domain, where $I$ denotes the input image, $y \in \{\mathbf{e}_i\mid\mathbf{e}_i=(0,0,\ldots,1,\ldots,0),1\leq i\leq x\} $ denotes generator-specific one-hot labels, $\mathbf{e}_i \in\mathbb{R}^{x}$ represents the $i$-th one-hot label, and $x$ is the number of generators. 

\subsection{Method Overview}

Unlike prior methods \cite{DNADet,clip} that focus on integrations within vision or vision-language modalities for DFA, our PVLM model pioneers the joint interactions of vision, language, and face parsing modalities to realize ZS-DFA. As Fig.~\ref{fig3} shows, our method includes three modules: multi-perspective visual encoder (MPVE), parsing encoder (PE), and language encoder (LE).

During training, given an input facial appearance image $I^\text{a}$, we leverage the Sobel operator \cite{sobel}, SRM \cite{SRM}, face parser \cite{faceparser}, and fine-grained text generator \cite{MFCLIP} to obtain the edge image $I^\text{e}$, noise image $I^\text{n}$, parsing image $P$, and fine-grained attribution prompts $T$, respectively. Subsequently, $I^\text{a}$, $I^\text{e}$, and $I^\text{n}$ are fed into the MPVE to capture global visual attribution embeddings $I_\text{v}$ across three views. Meanwhile, PVLM transmits $P$ and $T$ to PE and LE, to extract the diverse global parsing features $P_\text{g}$ and language attribution ones $T_\text{l}$, accordingly. Thereafter, we conduct the language-guided and parsing-guided contrastive learning, and flexible metric learning, respectively. Meanwhile, PVLM sends $I_\text{v}$ to the predictor and MLP, both consisting of fully connected layers, to yield the predicted language embeddings $T_\text{l}^{\text{pre}}$ and deepfake attribution ones, separately. Finally, we leverage the softmax function to produce the prediction $y_\text{pre}$.  

 During inference, to prevent information leakage, PVLM merely uses MPVE to dig visual features across three views, which are then imparted to the MLP head and softmax function, to yield the prediction.

\subsection{Multi-Perspective Visual Encoder}

To capture comprehensive and general visual attribution embeddings, we propose the MPVE module, since global representations are helpful for DFA as proven by \cite{DNADet}. Unlike existing multi-domain DFA methods \cite{MPSL,Freq} which tend to focus on the interaction between the local frequency and RGB information, our MPVE module can dig appearance global forgery traces, fine-grained noise global representations, and edge global features, and integrate them to explore diverse and common visual embeddings across three views.  As shown in Fig.~\ref{fig3}, MPVE mainly includes an appearance encoder (AE), an edge encoder (EE), a noise encoder (NE), and a multi-view multi-head interaction (MMI) module. 

To extract global appearance forgery features, we design AE, which consists of a backbone with stacked convolutional layers followed by a transformer encoder (TE) with several transformer blocks, following \cite{CViT}. Particularly, given an input face appearance image $I^\text{a}\in\mathbb{R}^{3\times H \times W}$, AE derives the local feature map $I^\text{a}_\text{loc} \in\mathbb{R}^{ c\times h\times w}$ via the backbone, where $H$ and $W$ denote the height and width of the image, and  $c$, $h$, $w$ are the channel, height, and width of the feature map, respectively. AE then sends it to the TE, to investigate global relationships among feature patches, to yield global appearance DFA embeddings $I^\text{a}_\text{g}\in\mathbb{R}^{2\times d}$, where $d$ denotes the dimension of the embeddings. AE obtains appearance class token $I^\text{a}_\text{cls}\in\mathbb{R}^{1\times d}$ and patch token $I^\text{a}_\text{p}\in\mathbb{R}^{1\times d}$ via disentangling $I^\text{a}_\text{g}$. Both EE and NE have similar architecture and workflow as AE. Given $I^\text{a}$, we employ the Sobel operator \cite{sobel} to acquire edge images $I^\text{e}\in\mathbb{R}^{1\times H \times W }$. We extract the richest image patch via the patch selector in \cite{MFCLIP} from $I^\text{a}$ and then send them to the SRM \cite{SRM} to gain the fine-grained noise image $I^\text{n}\in\mathbb{R}^{ 3\times p\times p}$, where $p$ denotes the size of the noise image. Like AE, EE and NE derive the global edge features $I^\text{e}_\text{g}\in\mathbb{R}^{2\times d}$ and noise embeddings $I^\text{n}_\text{g}\in\mathbb{R}^{ 2\times d}$, respectively.

{\bfseries\setlength\parindent{0em}Multi-view multi-head interaction module.} 
First, to enhance the efficient interaction between noise and appearance features, we employ multi-head cross-attention (MHCA) between $I_\text{g}^\text{n}$ and $I_\text{cls}^\text{a}$. Mathematically, the calculation of MHCA is as follows:

\begin{align}
	Q&={ I}_\text{cls}^\text{a}\ W_\text{q},\\
	K&=I_\text{g}^\text{n}\ W_\text{k},\\
	V&=I_\text{g}^\text{n}\ W_\text{v},
\end{align}
where $W_\text{q}, W_\text{k}, W_\text{v}\in\mathbb{R}^{d\times d_f}$, $f$ denotes the number of head space, and $d_f=d/f $ is the dimension of feature tokens in the head space. To achieve global diverse fusion across noise and appearance representations, we devise the formula as follows:

\begin{gather}
	\textrm{MHCA}({I}_\text{cls}^\text{a},I_\text{g}^\text{n})  =\textrm{softmax}(\frac{QK^T}{\sqrt{d_f}})V.
\end{gather}

Thereafter, the output of MHCA is projected and added to ${I}_\text{cls}^\text{a}$ to get ${I}_\text{cls}^\text{an}\in\mathbb{R}^{1\times d}$. 
The interaction between $I_\text{g}^\text{e}$ and $I_\text{cls}^\text{a}$ follows the same workflow as that between ${I}_\text{g}^\text{n}$ and ${I}_\text{cls}^\text{a}$. Therefore, we can gain diverse and rich fusion features ${I}_\text{cls}^\text{ae}\in\mathbb{R}^{1\times d}$ across appearance and edge views. Finally, we boost the communication between $I_\text{cls}^\text{an}$ and $I_\text{cls}^\text{ae}$ to yield visual attribution features across three views.

\begin{gather}
	I_\text{v} = I_\text{cls}^\text{an}\oplus I_\text{cls}^\text{ae},
\end{gather}
where $\oplus$ denotes the element-wise addition. $I_\text{v}$ is fed into the MLP head and then supervised by DFA loss to derive the generator-specific attribution features. (See Sect.~\ref{sec36})

\subsection{Parsing Encoder}
{\bfseries\setlength\parindent{0em}Motivation.} 
Different from current deepfake detection \cite{LampMark,Freq,DTN,tdsc,ADA-FInfer,Multi-View} or DFA methods \cite{MPSL,DNADet} that tend to introduce discriminative prior knowledge like identity, landmark, or frequency features, we focus on the face parsing information since the source face attributes in fake images tend to be incompletely preserved \cite{deepfakesurvey}. As Fig.~\ref{fig2} shows, considering that there are evident source face attributes preservation differences  among generators. Generally, GAN-generated face images retain more source face attribute information than diffusion-generated ones. 

{\bfseries\setlength\parindent{0em}Theoretical Analysis.}The generated facial appearance RGB image $I^\text{a}$ can be considered as a mixture of two potential signals: forgery-irrelevant semantic content signal $U_c$ and generator-specific forgery signal $U_f$. $U_c$ is determined by the face generation intention (e.g., identity, background), which has a large variance and is the main source of noise. $U_f$ is defined by the inherent properties of the generator $\mathcal{G}$ (e.g., architecture, hyperparameters), which has a weak amplitude and is the target signal. The observed signal is: $I^\text{a} = \mathcal{G}(U_c, U_f)$. Our deepfake attribution task aims to reliably estimate $U_f$ from $I^\text{a}$ to identify $\mathcal{G}$. In the original data space, the signal-to-noise ratio (SNR) of $U_f$ is extremely low, i.e., $ \text{SNR}_{\text{raw}} = \frac{\text{Var}(U_f)}{\text{Var}(U_c)} \approx 0 $. Traditional attribution methods directly learn the mapping $M(\mathcal{G} | I^a)$, which is prone to overfitting to the spurious statistical association between $U_c$ and $\mathcal{G}$ in the training set and forgery-unrelated semantic features that harm generalization across generators. To address this limitation, we design the parsing encoder (PE), which serves as a signal-to-noise ratio enhancer to capture generalizable generator-specific attribution features. 

PE is a function that takes a facial semantic parsing image $P$ generated by the pre-trained face parser \cite{faceparser} and outputs the face parsing-aware generator-specific forgery representations $P_g$, i.e., $P_g=\textrm{PE}(P)$. Unlike the RGB image, $P$ eliminates low-level features that carry rich identity information, such as texture and color, and preserves only the labels corresponding to facial semantic categories. The structure of $P$ prevents the PE from capturing the long-range identity-related features for identity bias learning, thereby actively disrupting the integrity of the semantic noise $U_c$. According to the data processing inequality \cite{mackay2003information}, forgery-irrelevant semantic content signals $U_c$, such as identity and background, in the output features are constrained by the limited information in the input face parsing images $P$, and the generator-specific forgery signals $U_f$ are thus enhanced. Besides, face parsing confines the face image to a specific semantic region, such as the lips. Within this subspace, the variability of $U_c$ is significantly reduced, as the shape variations of the lips are much smaller than the identity variations of the whole face, which forces our model to explore forgery pattern differences related to general generator-specific high-frequency manipulated details, and enhances the relative importance of $U_f$ in the local area. Therefore, PE implements a mapping from a high-noise observation space to a low-noise feature space, substantially improving the effective SNR of the generator-specific fingerprint signal:
\begin{align}
\text{SNR}_{\text{PE}} = \frac{\text{Var}(U_f \mid P)}{\text{Var}(U_c \mid P)} \gg \text{SNR}_{\text{raw}}, 
\end{align}
which facilitates our model to capture more general and discriminative forgery traces.
 
However, simply suppressing $U_c$ is insufficient to justify the effectiveness of PE. The key point is that generator-specific forgery signals are not completely independent of facial structure. We further decompose $U_f$ into a structure-dependent component $U_f^{str}$ and an appearance-only component $U_f^{app}$:
\begin{equation}
U_f = U_f^{str} \cup U_f^{app}.
\end{equation}
The structure-dependent component $U_f^{str}$ refers to generator-specific artifacts related to facial component layout, semantic boundaries, local shape deformation, region-wise inconsistency, and blending transitions. These cues are frequently produced by face synthesis and manipulation models because different generators exhibit different abilities in preserving facial attributes and semantic regions.

Compared with the appearance RGB image $I^\text{a}$, $P$ removes most fine-grained appearance information, such as texture, color, and background, which are strongly related to $U_c$. Therefore, according to the data processing inequality \cite{mackay2003information}, the parsing image constrains the information flow from forgery-irrelevant semantic content:
\begin{equation}
\mathrm{MI}(P;U_c) \leq \mathrm{MI}(I^\text{a};U_c),
\end{equation}
where $\mathrm{MI}(\cdot;\cdot)$ denotes mutual information~\cite{shannon1948mathematical}, measuring the statistical dependence between two variables. In our context, $\mathrm{MI}(P;U_c)$ quantifies the amount of forgery-irrelevant semantic information retained in the parsing image $P$.
More importantly, $P$ is not an arbitrary compression. It explicitly preserves facial component regions and their boundary configurations. Since $U_f^{str}$ is coupled with facial semantic layout and boundary structures, the parsing image still contains non-negligible generator-specific structural information:
\begin{equation}
\mathrm{MI}(P;U_f^{str}) > 0,
\end{equation}
where $\mathrm{MI}(P;U_f^{str})$ measures the amount of structure-dependent generator-specific forgery information preserved by $P$.
Thus, the PE does not aim to preserve all generator-specific signals. Instead, it preserves the structure-dependent component of $U_f$ while suppressing the identity or background-dominated component $U_c$.

Since PE is optimized by the dynamic cross-perspective contrastive (DCPC) loss to align the parsing-aware features $P_\text{g}$ generated by PE with the generator-specific visual attribution feature $I_\text{v}$, where $I_\text{v}$ is supervised by DFA loss to derive the generator-specific visual forgery features, the structure-dependent forgery information preserved in $P$ is further encouraged to be encoded into $P_\text{g}$. As a result, PE reduces the influence of $U_c$ while maintaining the structure-dependent generator-specific attribution cues:
\begin{equation}
	\frac{\mathrm{MI}(P_\text{g};U_f^{str})}{\mathrm{MI}(P_\text{g};U_c)}
	>
	\frac{\mathrm{MI}(I^\text{a};U_f^{str})}{\mathrm{MI}(I^\text{a};U_c)}.
\end{equation}
This indicates that the relative contribution of structure-dependent generator-specific forgery cues is increased in the parsing-guided representation $P_\text{g}$. Besides, as Fig.~\ref{figintro2} proves, the distributions of facial parsing images differ substantially between real and fake faces, and there are evident distribution differences in face parsing images derived from fake images generated by various generators. This phenomenon further facilitates the PE to preserve the parsing-aware generator-specific forgery traces under the guidance of the DCPC loss, which aims to encourage the positive-sample generator-discriminative visual forgery features $I_\text{v}$ to be closer to the PE-derived representation $P_\text{g}$. Therefore, PE can provide complementary attribution information to MPVE, rather than merely suppressing semantic content.

{\bfseries\setlength\parindent{0em}Framework.}
To align face parsing forgery features with multi-view visual ones and adapt face parsing information to our DFA domain, we design the parsing encoder (PE) consisting of a parsing backbone with stacked convolutional layers and a parsing transformer encoder (PTE) with $E$ transformer blocks, $\text{TB}_j^\text{p}$, $j=1,2,…, E$, to capture global and diverse face parsing forgery representations. As shown in Fig.~\ref{fig3}, given $I^\text{a}$, we use the CelebA-HQ pre-trained face parser \cite{faceparser} to generate the parsing image $P\in\mathbb{R}^{ 3\times H\times W}$. Considering that the face parser may not accurately extract the parsing information of unseen faces, due to the domain gap between the pre-trained face dataset and the unseen one, we propose to conduct the parsing-guided multi-view DFA representation learning only during training. Specifically, PE derives the local face parsing feature $P_\text{loc}\in\mathbb{R}^{ c\times h\times w}$ from $P$ through the backbone, to conduct the feature alignment. PE then imparts it to PTE to explore diverse long-range relations among features. In detail, $P_\text{loc}$ is flattened along the channel and projected to a 2D parsing token, and then appended with a learnable class token to capture the global parsing attribution features, to derive $P_\text{tok}=\text{App}(\text{Proj}(\text{Flat}(P_\text{loc})))\in\mathbb{R}^{ (n+1)\times d}$, where $n =  \frac{chw}{chw}$ denotes the number of the parsing token sequence and $d$ is the dimension of the parsing token. It is then added with learnable position representations $P_\text{e}\in\mathbb{R}^{(n+1)\times d}$ to encode the position information, i.e.,
\begin{align}
P_1^\text{tra}=P_\text{tok} \oplus P_\text{e}.
\end{align}
Subsequently, it is consecutively fed into $E$ blocks, i.e.,
\begin{align}
	\text{PTE}(p_1^{\text{tra}})
	&=\text{TB}^\text{p}_E\circ\text{TB}_{E-1}^\text{p}\circ \cdots \circ  \text{TB}_2^\text{p} \circ\text{TB}_1^\text{p}(P_1^\text{tra}) \nonumber \\
	&=\text{TB}_E^\text{p}\circ\text{TB}_{E-1}^\text{p}\circ \cdots \circ \text{TB}_2^\text{p}(P_2^\text{tra})
	\nonumber \\
	&= \cdots =\text{TB}_E^\text{p}\circ\text{TB}_{E-1}^\text{p}(P_{E-1}^\text{tra})\nonumber \\
	&= \text{TB}_E^\text{p}(P_E^\text{tra}) =p^\text{p}_\text{PTE},
\end{align}
where $\circ$ represents the function decomposition. PE derives global parsing-aware embeddings $P_\text{g}\in\mathbb{R}^{1\times d} $ using the class token in $ P^\text{p}_\text{PTE}$. To perform parsing-aware multi-view visual attribution representation learning and improve the generator-specific attribution features yielded by PE, we achieve the flexible vision-parsing matching via the proposed dynamic cross-perspective contrastive (DCPC) loss (See Sect.~\ref{sec36}), where the generator-specific visual forgery features $I_\text{v}$ of positive samples are encouraged to align with the representation $P_\text{g}$ derived by PE. This tends to boost the PE to preserve the parsing-aware generator-specific forgery features.

\subsection{Language Encoder}

To create fine and accurate text attribution prompts, we introduce the fine-grained text generator (FTG) \cite{MFCLIP}.  FTG generates text prompts for each image based on hierarchical fine-grained labels provided by the GenFace dataset. To prevent information leakage from texts and considering that unseen generators may lack text prompts, we introduce fine-grained texts to train our model only during training, to conduct language-guided multi-view representation learning. Specifically, given  $I^\text{a}$, we generate deepfake attribution prompts $T$ via FTG. To extract fine-grained language attribution-aware embeddings, we design the LE. As Fig.~\ref{fig3} shows, LE contains a language transformer encoder with $L$ transformer blocks, $\text{TB}_j^\text{l}$, $j=1,2,…, L$. In detail, given the fine-grained deepfake attribution prompts $T$, LE first yields a series of word tokens $T^\text{tok}\in\mathbb{R}^{t}$ using the tokenizer \cite{clip}, where $t$ represents the number of word tokens. They are then projected to the word embedding vector $T^\text{emb}\in\mathbb{R}^{t\times d}$, and added with learnable position embeddings $P_\text{l}\in\mathbb{R}^{t\times d}$, i.e., $T_\text{1}^\text{tra}={\ T}^\text{emb} \oplus P_\text{l}$. Thereafter, they are consecutively transferred to $L$ blocks, i.e., 
\begin{align}
	\text{LE}(T_\text{1}^{\text{tra}})
	&=\text{TB}_L^\text{l}\circ\text{TB}_{L-1}^\text{l}\circ \cdots \circ  \text{TB}_2^\text{l} \circ\text{TB}_1^\text{l}(T_1^\text{tra}) \nonumber \\
	&=\text{TB}_L^\text{l}\circ\text{TB}_{L-1}^\text{l}\circ \cdots \circ \text{TB}_2^\text{l}(T_2^\text{tra})
	\nonumber \\
	&= \cdots =\text{TB}_L^\text{l}\circ\text{TB}_{L-1}^\text{l}(T_{L-1}^\text{tra})\nonumber \\
	&= \text{TB}_L^\text{l}(T_L^\text{tra}) = T^\text{l}_\text{LE}.     
\end{align}
LE obtains the global deepfake attribution-perceptual language representations $T_\text{l}\in\mathbb{R}^{1\times d} $ using the class token in $T^\text{l}_\text{LE}$, which are then employed for language-vision matching (See Sect.~\ref{sec36}).
. 
\subsection{Loss Function}\label{sec36}

{\bfseries\setlength\parindent{0em} Deepfake attribution loss.}
To capture discriminative generator-specific DFA representations, we introduce the DFA loss as follows:
\begin{equation}
\mathcal{L}_\text{dfa}=\ \frac{1}{b}\sum_{u=1}^{b}{-({y^u})^T\text{log}(}y_\text{pre}^u),
\end{equation}
 where $y_\text{pre}\in\mathbb{R}^{x}$, $x$ is the number of the deepfake generators, $b$ denotes the number of samples in a batch, and $u\in b$ is the index of samples. 

{\bfseries\setlength\parindent{0em} Deepfake attribution contrastive center loss.}
However, learned DFA features are not general enough since the DFA loss merely considers the decision boundary among generators, rather than intra-generator compression and inter-generator separability. To address the limitation, we propose a novel DFACC loss. Unlike vanilla contrastive center (VCC) loss \cite{contrastive} which makes intra-generator samples closer and inter-generator ones farther apart, our DFACC loss minimizes the distance among within-generator samples, while placing correlated samples across generators closer and uncorrelated ones stay away from each other. In detail, our DFACC criterion is defined as:
\begin{equation}
	\mathcal{L}_\text{dfacc} = \mathcal{L}_\mathrm{intra}+\lambda \mathcal{L}_\mathrm{inter}.
\end{equation}

To achieve the intra-generator compaction, we design the $\mathcal{L}_\mathrm{intra}$ loss as follows:
\begin{equation}
	\mathcal{L}_\mathrm{intra}=\frac{1}{b}\sum_{u=1}^b\|y_\text{pre}^u-c_{y^u}\|_2,
\end{equation}
 where $c_{y^u} \in\mathbb{R}^{d}$ denotes the trainable center vector of the generator ${y^u}$, $\| \cdot \|_2$ is the Euclidean distance, $\lambda$ is the factor to balance the loss term, and $y^u$ denotes the generator-specific one-hot label of the sample $u$. To minimize the distance among relevant samples across generators, and maximize the distance among unrelated ones, we propose the $\mathcal{L}_\mathrm{inter}$ loss,

\begin{equation}
	\mathcal{L}_{\mathrm{inter}} = \frac{2}{b(b-1)}\sum_{u=1}^b\sum_{v=u+1}^b \mathbb{I}(y^u \neq y^v) \mathcal{L}_{uv},
\end{equation}
\vspace{-0.6em}
\begin{equation}
	\mathbb{I}(y^u \neq y^v) =
	\begin{cases}
		1, & \text{if } y^u \neq y^v, \\
		0, & \text{if } y^u = y^v,
	\end{cases}
\end{equation}
\vspace{-0.1em}
\begin{equation}
	\mathcal{L}_{uv}=
	\begin{cases}
		\max(\|y_\text{pre}^u-y_\text{pre}^v\|_2 -m, 0), & \hspace{-1em} \mathrm{if~}\|\mathbf{c}_{y^u}-\mathbf{c}_{y^v}\|_2<m, \\
		-\|y_\text{pre}^u-y_\text{pre}^v\|_2, & \hspace{-1em}\mathrm{if~}\|\mathbf{c}_{y^u}-\mathbf{c}_{y^v}\|_2\geq m,
	\end{cases}
\end{equation}
where $m$ is a predefined margin threshold to control the interval among samples, $\mathbb{I}$ is the indicator function, and $u, v=1,2,...,b$ is the sample index. 

{\bfseries\setlength\parindent{0em} Dynamic cross-perspective contrastive loss.}
To achieve the flexible vision-parsing matching, we devise the DCPC loss. Mathematically, for the vision-parsing feature pairs $\{(I_\text{v}^{u}, P_\text{g}^{u})\}_{u=1}^{b}$, we calculate the vision-parsing cosine similarity vector and the parsing-vision one as follows:

\begin{align}
	S_\text{v2p}^{uv}(I_\text{v}, P_\text{g})&=\frac{\text{exp}(\text{C}{(I}_\text{v}^u{,P}_\text{g}^v)/\tau)}{\sum_{v=1}^{b}{\text{exp}(\text{C}{(I}_\text{v}^u{,P}_\text{g}^v)/\tau)}}\odot\sigma(A^{uv}), \\
		S_\text{p2v}^{uv}(P_\text{g},I_\text{v} )&=\frac{\text{exp}(\text{C}{(P}_\text{g}^u{,I}_\text{v}^v)/\tau)}{\sum_{v=1}^{b}{\text{exp}(\text{C}{(P}_\text{g}^u{,I}_\text{v}^v)/\tau)}}\odot\sigma(A^{uv}),
\end{align}

where $\tau$ is a learnable temperature weight to smooth features, C (·) executes a dot product operation to compute similarity scores, $\sigma $ is a sigmoid function, $A\ \in\mathbb{R}^{b\times b}$ is a learnable weight matrix, $\odot$ is a element-wise product, and $u, v=1,2,...,b$ are the index of the sample. The one-hot label $y_\text{pa}$ of the $u$-th pair is denoted as $y_\text{pa}^u=\{{y_\text{pa}^{uv}}\}_{v=1}^b$, $y_\text{pa}^{uu}=1$,
$y_\text{pa}^{uv,u\neq v}=0$. To pull positive pairs together while pushing negative ones away, the DCPC loss is defined as $\mathcal{L}_\text{dcpc}=\ (\mathcal{L}_\text{v2p}+\mathcal{L}_\text{p2v})/2$, where $\mathcal{L}_\text{v2p}$ and $\mathcal{L}_\text{p2v}$ are denoted as follows:

\begin{align}
	\mathcal{L}_\text{v2p}
	&=\frac{1}{b}\sum_{u=1}^{b}{-({y_\text{pa}^u})^T\text{log}(}S_\text{v2p}^u),\\
	\mathcal{L}_\text{p2v}
	&=\frac{1}{b}\sum_{u=1}^{b}{-({y_\text{pa}^u})^T\text{log}(}S_\text{p2v}^u).
\end{align}
Like \cite{MFCLIP}, we introduce the cross-modal contrastive (CMC) loss to perform vision-language matching, and the Kullback-Leibler (KL) divergence loss to conduct the feature alignment. The total loss is as follows:
\begin{align}
	\mathcal{L}_\text{total} = 	\mathcal{L}_\text{dfa} +\mathcal{L}_\text{dfacc} +\mathcal{L}_\text{cmc}+ \mathcal{L}_\text{dcpc}+\mathcal{L}_\text{kl}.
\end{align}

\section{Experiments}

\begin{table}[t]
	\caption{The outline of the dataset construction and protocol. \# denotes the number of dataset. Pro means the protocol. \label{tabdata}}
	\setlength{\tabcolsep}{0.5mm}{
		\scriptsize
			\renewcommand{\arraystretch}{1.1} 
		\begin{tabular}{cccccccrr}
			\toprule
			Seen Sets & Unseen Sets &  Type & Sub-Type & Generator & Train \# & Test \# & \multicolumn{1}{l}{Pro-1} & \multicolumn{1}{l}{Pro-2} \\
			\midrule
			GenFace &  \multicolumn{1}{c}{-}     & EFS   & GAN   & StyleGAN2 & 10,000 & 1,250 &  \multicolumn{1}{c}{\checkmark}   &\multicolumn{1}{c}{\checkmark}  \\
			GenFace & \multicolumn{1}{c}{-}      & EFS   & GAN   & StyleGAN3 & 10,000 & 1,250 & \multicolumn{1}{c}{\checkmark}   &\multicolumn{1}{c}{\checkmark} \\
			GenFace &   \multicolumn{1}{c}{-}    & EFS   & Diffusion & LatDiff & 10,000 & 1,250 & \multicolumn{1}{c}{\checkmark}    &\multicolumn{1}{c}{\checkmark}  \\
			GenFace &   \multicolumn{1}{c}{-}    & EFS   & Diffusion & CollDiff & 10,000 & 1,250 & \multicolumn{1}{c}{\checkmark} & \multicolumn{1}{c}{\checkmark} \\
			GenFace &    \multicolumn{1}{c}{-}    & EFS   & Diffusion & DDPM  & 10,000 & 1,250 &  \multicolumn{1}{c}{\checkmark}     & \multicolumn{1}{c}{\checkmark} \\
			GenFace &   \multicolumn{1}{c}{-}     & FS    & GAN   & FSLSD & 10,000 & 1,250 & \multicolumn{1}{c}{\checkmark} &\multicolumn{1}{c}{\checkmark} \\
			GenFace &   \multicolumn{1}{c}{-}     & FS    & GAN   & FaceSwapper & 10,000 & 1,250 & \multicolumn{1}{c}{\checkmark} &\multicolumn{1}{c}{\checkmark}  \\
			GenFace &   \multicolumn{1}{c}{-}     & FS    & Diffusion & DiffFace & 10,000 & 1,250 & \multicolumn{1}{c}{\checkmark} &\multicolumn{1}{c}{\checkmark}  \\
			GenFace &   \multicolumn{1}{c}{-}     & AM    & GAN   & LatTrans & 10,000 & 1,250 &  \multicolumn{1}{c}{\checkmark}   &\multicolumn{1}{c}{\checkmark} \\
			GenFace &    \multicolumn{1}{c}{-}    & AM    & Diffusion & Diffae & 10,000 & 1,250 &  \multicolumn{1}{c}{\checkmark}    &\multicolumn{1}{c}{\checkmark}  \\
			\multicolumn{1}{c}{-}	& DF40  & EFS   & GAN   & VQGAN &    \multicolumn{1}{c}{-}   & 1,250 &  \multicolumn{1}{c}{\checkmark}   &\multicolumn{1}{c}{\checkmark} \\
			\multicolumn{1}{c}{-}	& DF40  & EFS   & Diffusion & SD-2.1 &  \multicolumn{1}{c}{-}     & 1,250 &  \multicolumn{1}{c}{\checkmark}   &\multicolumn{1}{c}{\checkmark}  \\
			\multicolumn{1}{c}{-}	& DF40  & EFS   & Diffusion & DiT   &  \multicolumn{1}{c}{-}     & 1,250 &  \multicolumn{1}{c}{\checkmark} &\multicolumn{1}{c}{\checkmark} \\
			\multicolumn{1}{c}{-}	& DF40  & EFS  & Diffusion & SiT  &  \multicolumn{1}{c}{-}     & 1,250 &  \multicolumn{1}{c}{\checkmark} &\multicolumn{1}{c}{\checkmark} \\
			\multicolumn{1}{c}{-}	& -  & EFS  & Flow & FLUX  &  \multicolumn{1}{c}{-}     & 1,250 &  \multicolumn{1}{c}{\checkmark} &\multicolumn{1}{c}{\checkmark} \\
			\multicolumn{1}{c}{-}	& DF40  & FS    & GAN   & SimSwap &    \multicolumn{1}{c}{-}   & 1,250 &  \multicolumn{1}{c}{\checkmark}   & \multicolumn{1}{c}{\checkmark} \\
			\multicolumn{1}{c}{-}	& DF40  & FS    & GAN   & UniFace & \multicolumn{1}{c}{-}      & 1,250 &   \multicolumn{1}{c}{\checkmark}    & \multicolumn{1}{c}{\checkmark} \\
			\multicolumn{1}{c}{-}	& DF40  & FS    & GAN   & InSwapper &    \multicolumn{1}{c}{-}   & 1,250 & \multicolumn{1}{c}{\checkmark}      & \multicolumn{1}{c}{\checkmark} \\
			\multicolumn{1}{c}{-}	& DF40  & FS    & GAN   & e4s   &    \multicolumn{1}{c}{-}     & 1,250 &   \multicolumn{1}{c}{\checkmark}    & \multicolumn{1}{c}{\checkmark} \\
			\multicolumn{1}{c}{-}	& \multicolumn{1}{c}{-}   & FS    & Diffusion & REFace &  \multicolumn{1}{c}{-}     & 1,250 &  \multicolumn{1}{c}{\checkmark}   & \multicolumn{1}{c}{\checkmark} \\
			\multicolumn{1}{c}{-}	& Celeb-DF++  & FR    & GAN   & LivePortrait  &    \multicolumn{1}{c}{-}     & 1,250 &   \multicolumn{1}{c}{\checkmark}    & \multicolumn{1}{c}{\checkmark} \\
			\multicolumn{1}{c}{-}	& Celeb-DF++ & FR    & Flow  & LIA  &    \multicolumn{1}{c}{-}     & 1,250 &   \multicolumn{1}{c}{\checkmark}    & \multicolumn{1}{c}{\checkmark} \\
			\multicolumn{1}{c}{-}	& Celeb-DF++  & FR   & GAN  & FSRT   &    \multicolumn{1}{c}{-}     & 1,250 &   \multicolumn{1}{c}{\checkmark}    & \multicolumn{1}{c}{\checkmark} \\
			\multicolumn{1}{c}{-}	& -  & FR   &Diffusion  & DiffusionAct   &    \multicolumn{1}{c}{-}     & 1,250 &   \multicolumn{1}{c}{\checkmark}    & \multicolumn{1}{c}{\checkmark} \\
			\multicolumn{1}{c}{-}	&Celeb-DF++  & TF    & GAN  & Real3D-Portrait   &    \multicolumn{1}{c}{-}     & 1,250 &   \multicolumn{1}{c}{\checkmark}    & \multicolumn{1}{c}{\checkmark} \\
			\multicolumn{1}{c}{-}	& Celeb-DF++  & TF   & Diffusion  & AniTalker   &    \multicolumn{1}{c}{-}     & 1,250 &   \multicolumn{1}{c}{\checkmark}    & \multicolumn{1}{c}{\checkmark} \\
			\multicolumn{1}{c}{-}	& Celeb-DF++ & TF    & Flow  &FLOAT   &    \multicolumn{1}{c}{-}     & 1,250 &   \multicolumn{1}{c}{\checkmark}    & \multicolumn{1}{c}{\checkmark} \\
			\multicolumn{1}{c}{-}	&Celeb-DF++  & TF   &Flow &EDTalk   &    \multicolumn{1}{c}{-}     & 1,250 &   \multicolumn{1}{c}{\checkmark}    & \multicolumn{1}{c}{\checkmark} \\

			CelebAHQ &    \multicolumn{1}{c}{-}    & Real  &     \multicolumn{1}{c}{-}   &   \multicolumn{1}{c}{-}     & 10,000 & 1,250 &   \multicolumn{1}{c}{$\times$}  & \multicolumn{1}{c}{\checkmark} \\
			\multicolumn{1}{c}{-}	& FFHQ  & Real  &     \multicolumn{1}{c}{-}   &     \multicolumn{1}{c}{-}   &   \multicolumn{1}{c}{-}     & 1,250 &    \multicolumn{1}{c}{$\times$}    &  \multicolumn{1}{c}{\checkmark}\\
			\bottomrule
		\end{tabular}%
	}
\end{table}

\subsection{Experiment Setup}

{\bfseries\setlength\parindent{0em} Dataset collection and protocol.}
We collect real and fake face images created by various advanced generators, including GAN, diffusion and flow model, based on several datasets, such as GenFace \cite{genface}, DF40 \cite{df40}, CelebAHQ \cite{celebahq}, Celeb-DF++ \cite{celebdf++}, and FFHQ \cite{ffhq}. As Table~\ref{tabdata} shows, these fake data mainly involve five manipulations, including entire face synthesis (EFS), attribute manipulation (AM), face reenactment (FR), talking face (TF), and face swap (FS). Each forgery contains some GAN-based, diffusion-based, and flow-based generators. We define two protocols to evaluate the attribution performance of various models under zero-shot scenarios: 1) \textbf{Protocol-1} intends to investigate the performance of different attributors on unseen generators. It consists of 28 generators, including GAN-based, diffusion-based, and flow-based ones from the GenFace, DF40, and Celeb-DF++ datasets. We generate 2k identity-swapped faces using the CelebA-HQ pre-trained REFace model \cite{refaces}, which is a diffusion-based FS method. We create 70k fake faces by inputting the real images from FFHQ to the FFHQ pre-trained DiffusionAct model \cite{refaces}, which is a diffusion-based FR method. We synthesize 2k entire manipulated faces using the flow-based FLUX model \cite{refaces}. 2) \textbf{Protocol-2} introduces additional original faces from both CelebA-HQ and FFHQ datasets based upon protocol-1. 

{\bfseries\setlength\parindent{0em} Evaluation metrics.}
We leverage accuracy (ACC) as our evaluation metric. For zero-shot settings, unlike previous methods \cite{MPSL, CPL} like OW-DFA that assign each novel deepfake to a specific category, considering that the categories of real-world generators are unknown and the goal of ZS-DFA is to identify generators beyond predefined categories, our approach groups all novel generators into a single unseen class. As shown in Fig.~\ref{fig3}, like \cite{hendrycks17baseline,ood}, we obtain the maximum confidence score from logits yielded by the softmax function, and define a threshold to evaluate whether the model has encountered the generator or not. We label the generator as unseen when the maximum falls below the threshold. The higher the confidence score, the more confident the deepfake attributor is that it has encountered the generator. The higher threshold means that the model exhibits greater confidence in correct classification. We set the threshold to 0.7 and 0.9, respectively, and only report results for the threshold of 0.9 on protocol-1 in the ablation study. We chose the threshold to 0.7 since it implies a fairly strong prediction but still leaves room for some uncertainty, which helps avoid being too strict and missing out on potentially useful classifications. We set the threshold to 0.9 because it denotes a very high level of confidence in the classification, and reduces the chance of misclassifications.

{\bfseries\setlength\parindent{0em}Implementation details.}
We implement the model using PyTorch on a Tesla V100 GPU with 32GB of memory under the Linux operating system. The number of transformer blocks $E$ and $L$ in PVLM are set to 6. The patch size $p$, feature dimension $d$, the factor $\lambda$, and batch size $b$ are set to 112, 512, 0.5, and 32, respectively. The number of transformer blocks in AE, EE, and NE are set to 6, 3, and 3, respectively. We set the channel $c$, height $h$, and width $w$ of feature maps to 512, 7, and 7, respectively. The height $H$ and width $W$ of images are configured to 224. We set the number $t$ of word tokens to 308. Our method is trained with the Adam optimizer \cite{adam} using a learning rate of 1e-4 and a weight decay of 1e-3. We leverage the scheduler to drop the learning rate by ten times every 15 epochs. We conduct the normalization by dividing the image pixel values by 255, without any data augmentation operations. To ensure a fair comparison, we develop the SOTA methods using the codes provided by the authors and follow the default configurations, until convergence. All models are trained and evaluated using the same dataset protocols.

\begin{table*}[t!]	
	\caption{ZS-DFA evaluation on protocol-1. ACC scores (\%) on unseen generators, after training using seen generators. ForAdapter is ForensicsAdapter \cite{forensics}. T means the threshold. \textbf{Bold} and \underline{underline} denote the best and the second-best results.\label{tabpro1}}
	\setlength{\tabcolsep}{1.00mm}{
		\scriptsize
	\renewcommand{\arraystretch}{1.3} 
		\begin{tabular}{|c|c|c|c|c|c|c|c|c|c|c|c|c|c|c|c|c|c|c|c|c|}
			\toprule
			\multirow{2}[4]{*}{Methods} & \multicolumn{2}{c|}{SimSwap} & \multicolumn{2}{c|}{InSwapper} & \multicolumn{2}{c|}{UniFace} & \multicolumn{2}{c|}{e4s} & \multicolumn{2}{c|}{VQGAN} & \multicolumn{2}{c|}{SD-2.1} & \multicolumn{2}{c|}{DiT} & \multicolumn{2}{c|}{SiT} & \multicolumn{2}{c|}{REFace} & \multicolumn{2}{c|}{Average} \\
			\cmidrule{2-21}      & T 0.7 &  T 0.9 & T 0.7 & T 0.9 & T 0.7 & T 0.9 & T 0.7 &T 0.9 & T 0.7 & T 0.9 & T 0.7 & T 0.9 & T 0.7 & T 0.9 & T 0.7 & T 0.9 & T 0.7 & T 0.9 & T 0.7 & T 0.9 \\
			\midrule
			CLIP \cite{clip}  &  \multicolumn{1}{c|}{9.12}  &  \multicolumn{1}{c|}{17.44} &  \multicolumn{1}{c|}{\underline{20.16}}  &  \multicolumn{1}{c|}{28.08}  &  \multicolumn{1}{c|}{18.24}  &  \multicolumn{1}{c|}{28.56} &   \multicolumn{1}{c|}{\textbf{37.60}} &  \multicolumn{1}{c|}{\underline{59.52 }} &  \multicolumn{1}{c|}{7.20} &  \multicolumn{1}{c|}{10.72}&  \multicolumn{1}{c|}{19.52}& \multicolumn{1}{c|}{ 36.32} & \multicolumn{1}{c|}{10.48}  &  \multicolumn{1}{c|}{17.76}  &  \multicolumn{1}{c|}{14.40}  & \multicolumn{1}{c|}{23.28}  &  \multicolumn{1}{c|}{\textbf{64.88}} & \multicolumn{1}{c|}{\textbf{77.76}} &  \multicolumn{1}{c|}{22.40} & \multicolumn{1}{c|}{33.27} \\
			CViT \cite{CViT} & 11.76  & 25.04 & 8.08  & 20.08  & 13.44  & 27.04 & 26.72 & 48.00  & 25.76  & 47.04 & 15.84 & 34.08 &21.52  & 39.52  & 20.24  &37.44  & 15.04 &31.92 & 17.60 &34.46 \\
			CAEL \cite{genface}  & 3.36  & 6.56  & 5.04  & 9.52  & 5.44  & 9.60  & 20.48  & 43.04  & 0.80  & 1.92  & \textbf{36.56} & \textbf{61.44} & 1.92  & 5.44  & 2.64  & 6.56  & 23.28 & 49.12 & 11.06 & 21.46 \\
			MFCLIP \cite{MFCLIP} & 21.36 & 46.72 & 13.76 & 31.12 & 20.40  & 44.96  & 29.76 & 57.04 & 14.88 & 35.76 & 12.72 & 26.16 & 13.52 & 30.16 & 10.64 & 27.12 & 25.92 & 49.28 & 18.11 & 38.70 \\
			DNA-Det \cite{DNADet}  & 6.04  & 16.72 & 4.20  & 10.2  & 3.04  & 10.16  & 28.76 & 59.28 & \underline{29.08} & 58.64 & 11.52 & 27.20  & 25.32 & 49.96 & 21.88  & 44.12 & 33.20  & 62.36  & 18.12  & 32.63  \\
			CPL \cite{CPL} & \underline{22.81} & \underline{67.90}  & 14.54& \underline{42.60} & \underline{22.67}  & \underline{56.82}  & \underline{30.48}  & \textbf{60.04}  & \textbf{30.80}  &  \underline{61.92}  & 13.74& 38.44 & \underline{26.92}  & \underline{65.40}  &\underline{ 22.70} &\underline{ 66.82}  & 35.28& 63.12 & \underline{26.68} & \underline{58.12} \\
			CARZero \cite{CARZero}   & \multicolumn{1}{c|}{ 2.21 } &  \multicolumn{1}{c|}{3.82}  &  \multicolumn{1}{c|}{2.07}  &  \multicolumn{1}{c|}{3.73}  & 2.43  & 3.71  & 9.36  & 17.04  & 3.61  & 6.92  & 7.91 & 17.44 & 2.03  & 3.59  & 3.01  & 5.37  & 9.36 & 20.05 & 4.67 & 9.07 \\
			DE-FAKE \cite{DEFAKE} & 0.48  & 1.36  & 0.48 & 1.52  & 0.96  & 2.64  & 7.12  & 15.36  & 2.00     & 5.52  & 6.96 & 15.76  & 0.72  & 2.24  & 1.28  & 3.04 & 8.24 &19.44 &  3.14     & 7.43 \\
			ForAdapter \cite{forensics} & 13.20  & 32.08  & 11.20 & 22.72  & 15.12 & 33.12  & 8.72  & 20.80  & 5.20    & 11.76  & 22.80 & 43.84 & 13.52  & 30.72  & 13.84 & 31.36&14.88&31.04&  13.16    & 28.60 \\
			PVLM (Ours)  &    \textbf{26.32}  &  \textbf{69.05}     &    \textbf{44.16}  &   \textbf{ 83.72 }  &    \textbf{26.05}   &   \textbf{66.98 }   &   12.41    &   36.72    &   26.59    &    \textbf{68.54 }  &  \underline{23.70}     &   \underline{ 47.02}   &   \textbf{ 42.94}   &  \textbf{ 80.72}    &   \textbf{41.79}    &   \textbf{81.13}    &    \underline{37.16}   &   \underline{64.02}   &    \textbf{31.24}  &   \textbf{66.43} \\
			\bottomrule
		\end{tabular}%
	}

\end{table*}

\vspace{-1em}
\subsection{Comparison with the State of the Art}

We evaluate state-of-the-art methods on our benchmark dataset. We
select various deepfake attributors such as hybrid transformer-based (i.e., CViT \cite{CViT} and CAEL \cite{genface}), vision-language-based (i.e., CLIP \cite{clip}, MFCLIP \cite{MFCLIP}, and ForensicsAdapter \cite{forensics}), zero-shot learning-based (CARZero \cite{CARZero}), and methods dedicated to attributing deepfake images (e.g., DNA-Det \cite{DNADet}, CPL \cite{CPL}, and DE-FAKE \cite{DEFAKE}).

{\bfseries\setlength\parindent{0em} Results on protocol-1.}
To study the attribution performance of different models, we conduct the ZS-DFA evaluation using protocol-1. We test models on various unseen generators from DF40 after training using seen generators from GenFace. In Table~\ref{tabpro1}, our method exceeds most attributors by a wide gap. It also surpasses the recent network MFCLIP by about 13\% ACC under the threshold of 0.7. Unlike MFCLIP which conducts the fine-grained language-guided image-noise forgery representation learning, our model introduces face parsing global features to conduct the bi-modality (e.g., text and parsing) guided multiple perspectives (e.g., image, noise, and edge) DFA representation learning. Compared to the DFA methods, the ACC of our method is nearly 4\% and 13\% higher than that of CPL and DNA-Det, validating the effectiveness of our model. Unlike DNA-Det, which only explores global attribution features within the image modality via patch-based contrastive learning, our model focuses on multiple modalities, including text and face parsing, as well as multiple views, such as image, noise, and edge, to achieve predominant ZS-DFA. Besides, the ACC score of models using the threshold of 0.9 is higher than that of models using the threshold of 0.7, which validates the reliability and effectiveness of the chosen threshold. We argue that a higher threshold necessitates that the model exhibit greater confidence in correct classification, thereby reducing the occurrence of misclassifications. 

{\bfseries\setlength\parindent{0em} Results on protocol-2.}
To further assess the attribution performance of various methods, we conduct the ZS-DFA evaluation on protocol-2. We test models on various unseen generators and FFHQ after training using both seen generators and CelebA-HQ. In Table~\ref{tabpro2}, the ACC of our method exceeds that of most SOTAs, showing the superior ZS-DFA capability of our method. Specifically, the average ACC score of our PVLM model is about 19.31\% higher than that of CViT on protocol-2. Unlike CViT merely explores global image forgery traces, our method focuses on global visual forgery features across three views (i.e., image, noise, and edge) and conducts fine-grained vision-language alignment and vision-parsing matching. We also test models on various unseen generators, including LivePortrait, FSRT, LIA, DiffusionAct, AniTalker, Real3D-Portrait, FLOAT, EDTalk, and FLUX, after training using both seen generators and CelebA-HQ. In Table~\ref{generator}, the average ACC score of our PVLM model is about 21.08\% higher than that of MFCLIP. Different from MFCLIP which merely mines global fine-grained visual forgery patterns across noise and image modalities and enhances them via vision-language matching, our PVLM method further introduces the distinct face parsing priors and captures comprehensive and complementary forged traces across image, noise, and face parsing views via vision-language alignment and flexible vision-parsing matching. Besides, models trained using protocol-2 outperform those trained with protocol-1. For instance, MFCLP and DNA-Det gain the average ACC score of 38.70\% and 32.63\% under the threshold of 0.9 setting on protocol-1, respectively. By contrast, they attain 43.38\% and 50.48\% ACC on protocol-2. It is noticed that a nearly 4.7\% and 17.85\% growth of ACC could be realized by adding original face distributions, which shows that pristine face images could facilitate the ZS-DFA performance of various deepfake attributors. We argue that models may extract more discriminative and common forgery features for each generator, due to the addition of authentic face distributions.

{\bfseries\setlength\parindent{0em} Generalization to unseen in-the-wild deepfake datasets.}
To offer stronger evidence of domain robustness beyond synthetic benchmarks, we train networks using protocol-2 and test them on unseen in-the-wild deepfake datasets, such as WildDeepfake \cite{wilddeepfake}, Celeb-DF \cite{Celeb-DF}, and DFDC \cite{DFDC}. In Table~\ref{tabpro3}, our model surpasses most methods across various fake face images from unseen datasets, which shows that our model achieves excellent domain robustness. To conduct the runtime benchmark, we assess the computational costs of models, including the parameter count and inference time. Although our PVLM model has a larger number of parameters than the lightweight DNA-Det, the attribution performance of PVLM is significantly improved on unseen deepfake datasets. In particular, the ACC score of PVLM is about 30\% higher than that of DNA-Det, with an increase of 236.02M parameters but a reduction of 0.02s in the inference time. In the future, we aim to apply knowledge distillation and model architecture modifications, such as a window-based multi-head self-attention, to achieve faster efficiency. 

\begin{figure*}[t!]
	\centering
	\includegraphics[width=\linewidth]{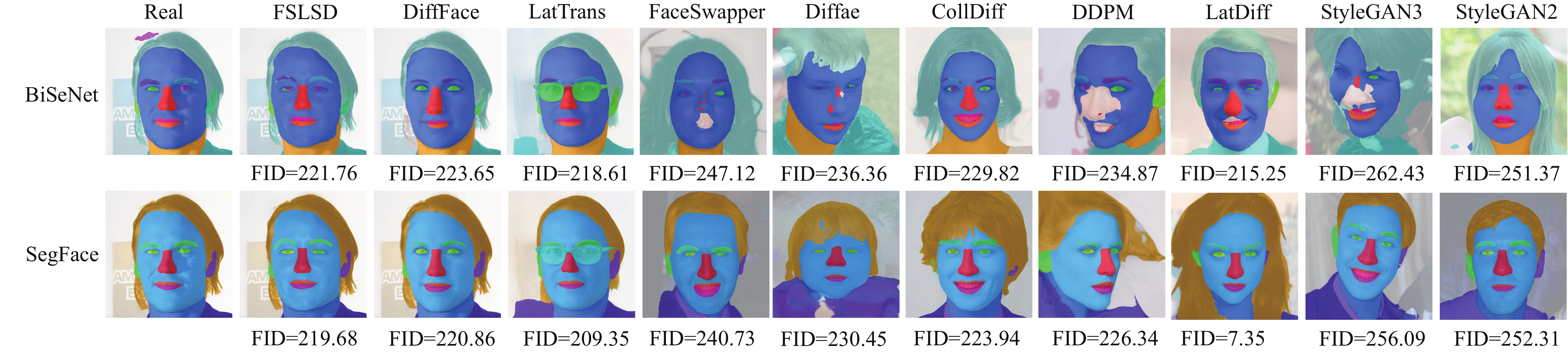} 
	\caption{ Visualization of parsing images produced by different face parsers for fake images synthesized by various generators.}\label{fig122}
\end{figure*}

\begin{table*}[t]
	
	\caption{ZS-DFA evaluation on protocol-2. ACC scores (\%) on unseen generators and FFHQ, after training using seen generators and CelebA-HQ. T means the threshold. \textbf{Bold} and \underline{underline} denote the best and the second-best results. \label{tabpro2}}
	\setlength{\tabcolsep}{0.48mm}{
		\scriptsize
		\renewcommand{\arraystretch}{1.3} 
		\begin{tabular}{|c|c|c|c|c|c|c|c|c|c|c|c|c|c|c|c|c|c|c|c|c|c|c|}
			\toprule
			\multirow{2}[4]{*}{Methods} & \multicolumn{2}{c|}{SimSwap} & \multicolumn{2}{c|}{InSwapper} & \multicolumn{2}{c|}{UniFace} & \multicolumn{2}{c|}{e4s} & \multicolumn{2}{c|}{VQGAN} & \multicolumn{2}{c|}{SD-2.1} & \multicolumn{2}{c|}{DiT} & \multicolumn{2}{c|}{SiT} & \multicolumn{2}{c|}{REFace} & \multicolumn{2}{c|}{FFHQ} & \multicolumn{2}{c|}{Average} \\
			\cmidrule{2-23}      & T 0.7 & T 0.9 & T 0.7 & T 0.9 & T 0.7 & T 0.9 & T 0.7 & T 0.9 & T 0.7 & T 0.9 & T 0.7 & T 0.9 &  T 0.7 &  T 0.9 &  T 0.7 & T 0.9 &  T 0.7 &  T 0.9 & T 0.7 &  T 0.9 &  T 0.7 & T 0.9 \\
			\midrule
			CLIP \cite{clip} &16.80 & 21.92& 19.12 & 27.04 & 14.88  & 21.20& \textbf{40.56}  &  62.80 & 12.16 & 15.36 & 24.64& 42.24& 14.08& 18.72 &15.92& 23.04  & \textbf{86.80}&\textbf{88.64}& \textbf{25.84}     &  \textbf{38.80} &   27.08    &  35.98\\
			CViT \cite{CViT} & 28.24  & 49.04 & 19.44  & 39.04  & 29.84  & 50.00 & 21.20  & 37.04  & 16.16 & 31.52 & 17.60 & 42.40& 22.24  & 45.68  &  \underline{22.96}  & \underline{46.56}  & 3.36  &8.32 & 2.56     &  7.52    &  18.36     & 35.71\\
			CAEL \cite{genface}    & 4.88  & 12.24 & 8.96  & 21.6  & 6.96  & 18.00  & 33.12 & \underline{65.36}& 18.40  & 36.64 & \textbf{31.92} & \textbf{64.80} & 3.52  & 8.08  & 4.48  & 9.60   & 30.40  & 62.08 & 3.92      &   10.16    &  14.66     & 30.86 \\
			MFCLIP \cite{MFCLIP} & 33.68 & 63.20  & 35.28 & 64.16 & 34.08  & 63.84  & 24.96 & 50.48 & \underline{34.72} & 63.12 & 11.76 & 25.84 & 22.00    & 39.28 & 18.88 & 33.84 & 10.48 & 28.32 & 0.88  & 1.76  & 22.67 & 43.38 \\
			DNA-Det \cite{DNADet}  & 27.56 & 61.60  & 26.28 & 63.84  & 18.20  & 50.16  & \underline{ 39.48}  &\textbf{77.56 } &  26.32 & 50.00    & 19.84  & 34.08  & 16.68  & 39.32  & 13.20  & 33.16  & 42.32  &79.16 & 7.96  & 15.96 & 23.78 & 50.48 \\
			CPL \cite{CPL}  &     \underline{35.70}   &  \underline{68.38}   &   \underline{ 37.03 }   &   \underline{ 75.77}    &  \underline{36.26} &     \textbf{75.49}  &    28.55   & 62.06     & \textbf{35.69}     &  \textbf{74.58}    &   12.65    &   37.99    &    \underline{24.87} &  \underline{51.53}    &  20.21  &  45.60   &    42.72  &    \underline{80.51}   &   8.90    &    16.64   &   \underline{28.26}& \underline{58.86}\\
			CARZero \cite{CARZero}  &    3.04   &    4.90   &   2.76    &    3.49   &    2.54   & 3.09 &   10.43  &    20.44   &     3.73  &    7.21   &  6.19     &   14.03    &    1.36   &   3.45    &    2.88   &  5.25     &  9.37     &   19.02    &  13.47     &   25.01   &       6.20     & 11.72 \\
			DE-FAKE \cite{DEFAKE} & 1.12  & 2.32  & 1.36  & 2.88  & 1.20  & 2.48  & 8.56  & 19.84  & 2.40   & 5.12  & 5.92  & 13.28 & 0.80   & 1.92  & 1.60   & 3.52  & 6.48  & 17.44  & 11.68 & 24.24 & 4.11  & 9.30 \\
			ForAdapter \cite{forensics} & 12.48  & 30.00  & 16.88  & 33.60  & 15.04  & 33.68  & 21.60  & 45.20  & 1.76  & 5.04  & 16.56  & 34.48 & 15.92  & 32.80  & 16.24   & 35.76  & 24.40  & 48.16  & 5.44 & 12.48 & 14.63  & 31.12 \\
			PVLM (Ours)  &      \textbf{ 36.89}  &   \textbf{72.05}    &   \textbf{54.71 }   &    \textbf{87.62}   & \textbf{39.07}      &     \underline{ 72.53} &      17.95 &    40.84   &  34.51    &  \underline{ 73.68 }    &   \underline{ 30.91 }   &     \underline{ 49.46 } & \textbf{ 50.77}     &  \textbf{84.89  }   &\textbf{ 50.43}      &  \textbf{88.62}     &  \underline{45.68 }    & 79.03   &\underline{15.80}     &\underline{27.35} &\textbf{37.67} &\textbf{67.61}
			\\
			\bottomrule
		\end{tabular}%
	}
\end{table*}

\begin{table*}[t!]	
	\caption{ZS-DFA evaluation on protocol-2. ACC scores (\%) on unseen generators, after training using seen generators and CelebAHQ. ForAdapter is ForensicsAdapter \cite{forensics}. T means the threshold. \textbf{Bold} and \underline{underline} denote the best and the second-best results.\label{generator}}
	\setlength{\tabcolsep}{0.98mm}{
		\scriptsize
		\renewcommand{\arraystretch}{1.3} 
		\begin{tabular}{|c|c|c|c|c|c|c|c|c|r|r|r|r|r|r|r|r|r|r|r|r|}
			\toprule
			\multirow{2}[4]{*}{Methods} & \multicolumn{2}{c|}{LivePortrait} & \multicolumn{2}{c|}{LIA} & \multicolumn{2}{c|}{FSRT} & \multicolumn{2}{c|}{DiffusionAct} & \multicolumn{2}{c|}{EDTalk} & \multicolumn{2}{c|}{AniTalker} & \multicolumn{2}{c|}{Real3DPortrait} & \multicolumn{2}{c|}{FLOAT} & \multicolumn{2}{c|}{FLUX} & \multicolumn{2}{c|}{Average} \\
			\cmidrule{2-21}      & T 0.7 & T  0.9 & T  0.7 & T  0.9 & T  0.7 & T  0.9 & T  0.7 & T 0.9 & \multicolumn{1}{c|}{T 0.7} & \multicolumn{1}{c|}{T 0.9} & \multicolumn{1}{c|}{T 0.7} & \multicolumn{1}{c|}{T 0.9} & \multicolumn{1}{c|}{T 0.7} & \multicolumn{1}{c|}{T 0.9} & \multicolumn{1}{c|}{T 0.7} & \multicolumn{1}{c|}{T 0.9} & \multicolumn{1}{c|}{T 0.7} & \multicolumn{1}{c|}{T 0.9} & \multicolumn{1}{c|}{T 0.7} & \multicolumn{1}{c|}{T 0.9} \\
			\midrule
			CLIP   \cite{clip}& 22.72 & 33.60  & 22.40  & 36.88 & 22.72 & 37.52 & 35.68 & 53.84 & \multicolumn{1}{c|}{11.68} & \multicolumn{1}{c|}{24.48} & \multicolumn{1}{c|}{13.04} & \multicolumn{1}{c|}{27.36} & \multicolumn{1}{c|}{6.88} & \multicolumn{1}{c|}{15.92} & \multicolumn{1}{c|}{4.96} & \multicolumn{1}{c|}{14.16} & \multicolumn{1}{c|}{3.60} & \multicolumn{1}{c|}{8.16} & \multicolumn{1}{c|}{15.96} & \multicolumn{1}{c|}{27.99} \\
			CViT  \cite{CViT} & 9.60   & 17.04 & 3.12  & 7.44  & 2.00     & 6.96  & 9.04  & 22.32 & \multicolumn{1}{c|}{8.80} & \multicolumn{1}{c|}{18.72} & \multicolumn{1}{c|}{0.72} & \multicolumn{1}{c|}{1.84} & \multicolumn{1}{c|}{25.76} & \multicolumn{1}{c|}{54.64} & \multicolumn{1}{c|}{33.52} & \multicolumn{1}{c|}{58.00} & \multicolumn{1}{c|}{\underline{13.60}} & \multicolumn{1}{c|}{\underline{30.08}} & \multicolumn{1}{c|}{11.80} & \multicolumn{1}{c|}{24.12} \\
			CAEL \cite{genface} & 21.36 & 43.44 & 31.68 & 60.32 & 29.04 & 58.08  & 8.96  & 21.92  & \multicolumn{1}{c|}{27.04} & \multicolumn{1}{c|}{54.72} & \multicolumn{1}{c|}{25.76} & \multicolumn{1}{c|}{51.04} & \multicolumn{1}{c|}{21.68} & \multicolumn{1}{c|}{44.80} & \multicolumn{1}{c|}{5.84} & \multicolumn{1}{c|}{15.84} & \multicolumn{1}{c|}{6.40} & \multicolumn{1}{c|}{15.12} & \multicolumn{1}{c|}{19.75} & \multicolumn{1}{c|}{40.59} \\
			MFCLIP  \cite{MFCLIP} & \underline{42.08} & \underline{74.88} & \underline{44.24} &\underline{81.52} & \underline{43.76}  & \underline{81.92}  & 2.32  & 6.72  & \multicolumn{1}{c|}{42.8} & \multicolumn{1}{c|}{\underline{83.28}} & \multicolumn{1}{c|}{37.68} & \multicolumn{1}{c|}{\underline{69.52}} & \multicolumn{1}{c|}{28.56} & \multicolumn{1}{c|}{60.88} & \multicolumn{1}{c|}{\underline{43.76}} & \multicolumn{1}{c|}{72.32} & \multicolumn{1}{c|}{6.08} & \multicolumn{1}{c|}{12.40} & \multicolumn{1}{c|}{\underline{32.36}} & \multicolumn{1}{c|}{\underline{60.38}} \\
			DNA-Det \cite{DNADet} & 1.92  & 4.24  & 0.00     & 0.32  & 0.16  & 0.64  & \underline{47.52}  & \underline{75.36}  & \multicolumn{1}{c|}{0.00} & \multicolumn{1}{c|}{0.02} & \multicolumn{1}{c|}{0.00} & \multicolumn{1}{c|}{0.08} & \multicolumn{1}{c|}{29.36} & \multicolumn{1}{c|}{71.44} & \multicolumn{1}{c|}{41.28} & \multicolumn{1}{c|}{\underline{74.56}} & \multicolumn{1}{c|}{8.64} & \multicolumn{1}{c|}{17.76} & \multicolumn{1}{c|}{14.32} & \multicolumn{1}{c|}{27.16} \\
			CPL \cite{CPL}   &   \multicolumn{1}{c|}{19.56 }    &  \multicolumn{1}{c|}{ 42.71}     &  \multicolumn{1}{c|}{ 30.79 }    &  \multicolumn{1}{c|}{ 70.54 }    &  \multicolumn{1}{c|}{ 40.02 }    &   \multicolumn{1}{c|}{ 78.95 }   &  \multicolumn{1}{c|}{21.04}      &    \multicolumn{1}{c|}{ 70.85}   &   \multicolumn{1}{c|}{ \underline{30.41} }   &   \multicolumn{1}{c|}{ 60.79 }   &   \multicolumn{1}{c|}{ \underline{40.13} }   & \multicolumn{1}{c|}{ 53.79 }     &     \multicolumn{1}{c|}{\underline{30.46}}  &  \multicolumn{1}{c|}{ \underline{72.88} }    &   \multicolumn{1}{c|}{ 40.37 }   &   \multicolumn{1}{c|}{ 70.84 }   &    \multicolumn{1}{c|}{ 7.20}   & \multicolumn{1}{c|}{ 13.97 }     &   \multicolumn{1}{c|}{28.89 }   & \multicolumn{1}{c|}{ 59.48 } \\
			CARZero \cite{CARZero}&  \multicolumn{1}{c|}{ 5.89}    & \multicolumn{1}{c|}{12.05}      &  \multicolumn{1}{c|}{  6.77}   & \multicolumn{1}{c|}{ 16.97 }     &  \multicolumn{1}{c|}{  8.02}   &   \multicolumn{1}{c|}{ 18.56}   &   \multicolumn{1}{c|}{ 12.76}    &   \multicolumn{1}{c|}{20.18}      &     \multicolumn{1}{c|}{ 6.89}   &   \multicolumn{1}{c|}{  10.97  }  &    \multicolumn{1}{c|}{ 7.54 }   &    \multicolumn{1}{c|}{  14.37 }  &   \multicolumn{1}{c|}{ 5.09 }    &   \multicolumn{1}{c|}{ 11.44}     &    \multicolumn{1}{c|}{ 8.96 }   &    \multicolumn{1}{c|}{ 15.73 }   &   \multicolumn{1}{c|}{  13.89 }   &    \multicolumn{1}{c|}{ 29.40 }   &  \multicolumn{1}{c|}{8.42}     & \multicolumn{1}{c|}{16.65} \\
			DE-FAKE \cite{DEFAKE}&    4.40   & 11.04      &    5.84   &   14.48    &  6.96     &  17.60     &  10.48     &    23.84   &   \multicolumn{1}{c|}{ 5.28 }  &  \multicolumn{1}{c|}{ 12.16 }   & \multicolumn{1}{c|}{ 5.92 }    & \multicolumn{1}{c|}{ 13.52 }    &   \multicolumn{1}{c|}{ 3.60 }  &  \multicolumn{1}{c|}{ 9.60 }   & \multicolumn{1}{c|}{  6.08 }   &  \multicolumn{1}{c|}{ 14.16 }   &  \multicolumn{1}{c|}{12.00 }    &  \multicolumn{1}{c|}{ 28.32 }   &  \multicolumn{1}{c|}{ 6.73 }     & \multicolumn{1}{c|}{ 16.08 } \\
			ForAdapter \cite{forensics} & 33.44 & 64.16 & 26.00    & 52.88 & 26.08  & 55.44  & 23.76 & 47.20  & \multicolumn{1}{c|}{28.96} & \multicolumn{1}{c|}{56.32} & \multicolumn{1}{c|}{29.44} & \multicolumn{1}{c|}{56.56} & \multicolumn{1}{c|}{21.92} & \multicolumn{1}{c|}{48.32} & \multicolumn{1}{c|}{33.52} & \multicolumn{1}{c|}{58.88} & \multicolumn{1}{c|}{10.08} & \multicolumn{1}{c|}{22.48} & \multicolumn{1}{c|}{25.91} & \multicolumn{1}{c|}{51.36} \\
			PVLM (Ours)   & \multicolumn{1}{c|}{ \textbf{46.88} }    &  \multicolumn{1}{c|}{\textbf{ 78.08} }    &  \textbf{50.40}     &  \textbf{82.56 }    &  \textbf{52.08}     &  \textbf{ 86.88 }   &     \multicolumn{1}{c|}{\textbf{61.76}}  &  \multicolumn{1}{c|}{ \textbf{83.60}}    &  \multicolumn{1}{c|}{\textbf{56.24}}     & \multicolumn{1}{c|}{\textbf{ 88.40 }}    &    \multicolumn{1}{c|}{\textbf{49.12}}    &   \multicolumn{1}{c|}{\textbf{74.88}}       &  \multicolumn{1}{c|}{ \textbf{61.84} }  &    \multicolumn{1}{c|}{ \textbf{92.32 }}   &  \multicolumn{1}{c|}{\textbf{ 61.92}  }  &  \multicolumn{1}{c|}{ \textbf{88.48} }  &  \multicolumn{1}{c|}{ \textbf{32.00} }   &  \multicolumn{1}{c|}{\textbf{ 57.92} }    &  \multicolumn{1}{c|}{ \textbf{52.47 }}    & \multicolumn{1}{c|}{\textbf{81.46}} \\
			\bottomrule
		\end{tabular}%
		
	}

\end{table*}

\begin{table}[t!]
	\caption{ZS-DFA evaluation on unseen in-the-wild deepfake datasets. T is the threshold. ACC scores (\%) on unseen deepfake datasets, after training using seen generators and CelebA-HQ.\label{tabpro3}}
	\setlength{\tabcolsep}{0.35mm}{
		\scriptsize
			\renewcommand{\arraystretch}{1.3} 
		
		\begin{tabular}{|c|c|c|c|c|c|c|c|c|c|c|}
			\toprule
			\multirow{2}[4]{*}{Methods} & \multicolumn{2}{c|}{Celeb-DF} & \multicolumn{2}{c|}{DFDC} & \multicolumn{2}{c|}{WildDeepfake} & \multicolumn{2}{c|}{Average} & \multirow{2}[4]{*}{\shortstack{Params  \\(M)}} & \multirow{2}[4]{*}{\shortstack{Inference \\Time (s)}} \\
			\cmidrule{2-9}      & T 0.7 & T 0.9 & T 0.7 & T 0.9 & T  0.7 & T  0.9 & T 0.7 & T  0.9 &       &  \\
			\midrule
			CLIP \cite{clip}  & 2.56  & 8.56  & 8.64  & 17.92 & 17.68 & 39.52 & 9.63  & 22.00    & 84.23 & 1.41 \\
			CViT \cite{CViT}& 7.28  & 14.24 & 25.84 & 47.36 & 19.84 & 43.20  & 17.65 & 34.93 & 89.02 & 0.26\\
			CAEL \cite{genface} & 12.96 & 28.32 & 9.52  & 23.44 & 21.76 & 49.52  & 14.75  & 33.76  & 158.63  &3.74  \\
			MFCLIP \cite{MFCLIP} & 34.88 & 62.32 & \underline{28.48} & \underline{58.16} & \underline{ 29.68}  & \underline{60.64}  & \underline{31.01} & \underline{60.37} & 93.83 & 1.67 \\
			DNA-Det \cite{DNADet} & 17.12 & 41.36 & 25.04 & 48.08  & 0.90  & 3.28  & 14.35  & 30.91  & 6.23  & 0.13 \\
			CPL \cite{CPL} &     \underline{35.06}  &   \underline{64.24}    &  \textbf{29.50}     &  57.32     &    24.91   & 56.02      &  29.82     &     59.19  &     10.79  &0.17 \\
			CARZero \cite{CARZero} &  7.86     &  12.65     &      8.09 &   10.64    &    9.37   &    25.90 &  8.44      &   16.36   &  84.24     & 1.42 \\
			DE-FAKE \cite{DEFAKE} & 4.16  & 9.92  & 2.08  & 4.4   & 7.04  & 14.64  & 4.43  & 9.65  & 84.24 & 1.42 \\
			ForAdapter\cite{forensics} & 11.36  & 27.60  & 20.48  & 47.44   &25.36 & 56.08  & 19.07  & 43.71  & 5.70 & 0.57 \\ 
		PVLM (Ours)   &    \textbf{36.14}  &  \textbf{68.24}     &    27.78   &   \textbf{60.75}   &   \textbf{31.28}    &  \textbf{68.56}     &   \textbf{31.73}    &   \textbf{65.85}    & 242.25 & 0.11 \\
			\bottomrule
		\end{tabular}%
	}
\end{table}

\begin{figure*}[t]
	\centering
	\includegraphics[width=0.95\linewidth]{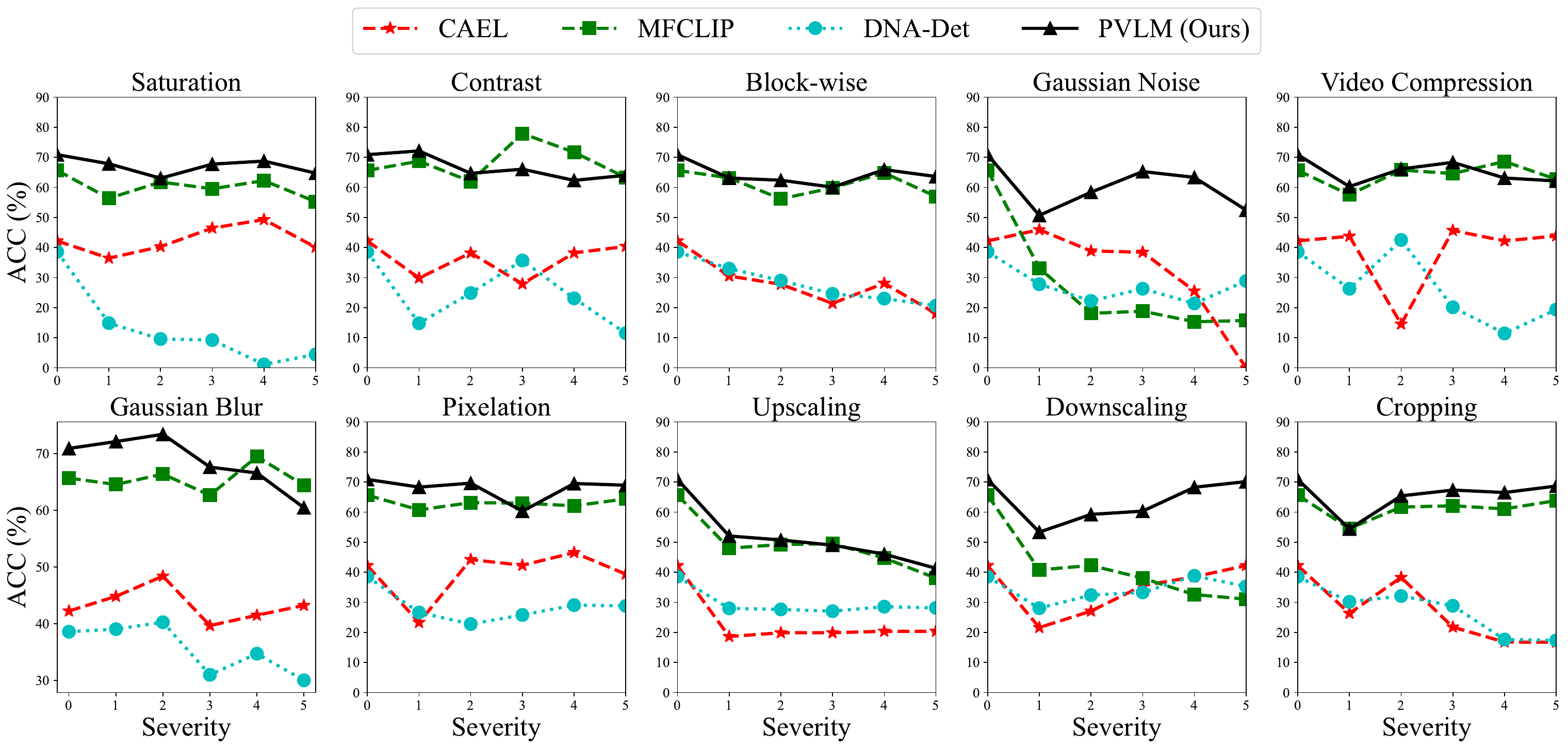} 
	\caption{Robustness to unknown image deformations.}\label{fig19}
\end{figure*}

{\bfseries\setlength\parindent{0em} Robustness to unseen image corruptions.} We evaluate the robustness of deepfake attributors against various unseen image perturbations. We train networks using protocol-2, and test their ZS-DFA performance on unseen distorted images from \cite{df10}. Ten types of corruption are engaged, each with five severity levels. As Fig.~\ref{fig19} displays, we test attributors on diverse image distortions, including saturation changes, contrast adjustments, block distortions, white Gaussian noise, blurring, pixelation, video compression, upscaling, downscaling, and cropping. A severity of 0 means no corruption. Specifically, the specific settings for the severity of the seven deformations are discussed in detail in \cite{MFCLIP}. In Fig.~\ref{fig19}, we notice that
the ACC of networks tend to decrease with the growth of
downscaling severities. By contrast, the ACC of our PVLM
method shows a rising trend, verifying the
strong attribution robustness of PVLM. Furthermore, the results demonstrate that our model outperforms most existing methods across different types of image degradation.

\subsection{Ablation Study}
We conduct ablation study including impacts of various modules, the face parser, various losses, the fine-grained text prompts, the MMI module, the margin or factor in DFACC loss.

{\bfseries\setlength\parindent{0em} Effects of various modules.}
To evaluate the contribution of each module to learning ability, we observe the DFA performance on both seen and unseen generators using protocol 1. In Table~\ref{tabab}, NE improves the performance of the attributor by 2.34\% ACC on images created by unseen generators, showing that SRM noises provide significant clues to facilitate ZS-DFA. The improvement from incorporating EE (+2.48\%) is clear, highlighting the importance of edge global embeddings. LE further boosts performance by about 21\% ACC, when EE or PE is also involved. The gain brought by LE alone is marginal without the integration of EE or PE, as the effectiveness of LE relies on the synergy between the different features. LE provides fine-grained semantic guidance, which is enhanced when combined with the distinctive facial parsing prior information from PE or edge features from EE. Without these modules, LE's potential is not fully realized, leading to a smaller performance gain. Besides, when the MPVE module (i.e. AE+NE+EE) is involved, the ACC of the model is 43.67\% on unseen generators. The improvement from adding PE (+15.36\%) is evident, highlighting the importance of discriminative facial parsing guidance. LE enhances performance by about 7.4\% ACC on unseen generators. The combined gain from integrating both PE and LE (+22.76\%) is particularly striking. We argue that fine-grained language embeddings from LE and facial attribute information from PE reinforce and enhance each other, synergistically guiding the model to capture both general and distinct attribution embeddings.

\begin{table}[t]
	\begin{minipage}{0.46\linewidth}
		\centering		
				\caption{Effects of different contrastive center losses.}	\label{abccloss}
		\setlength{\tabcolsep}{1.80mm}{
			\scriptsize
				\renewcommand{\arraystretch}{1.3} 
			\begin{tabular}{lcc}
				\toprule
				\multicolumn{1}{c}{\multirow{3}[6]{*}{Method}} & \multicolumn{2}{c}{Protocol-1} \\
				\cmidrule{2-3}      & Seen  & Unseen \\
				\cmidrule{2-3}      & ACC   & ACC \\
				\midrule
				CLIP w/VCC & 79.04 &-  \\
				CLIP w/DFACC & 89.07 &-  \\
				\midrule
				Ours w/VCC & 96.90 & 62.88 \\
				Ours w/DFACC & \textbf{100}   &\textbf{ 66.43} \\
				\bottomrule
			\end{tabular}%
		}

	\end{minipage}
	\hfill
	\begin{minipage}{0.48\linewidth}  
		\centering
				\caption{Effects of various face parsers.\label{tabfp}}
		\setlength{\tabcolsep}{0.7mm}{
			\scriptsize
				\renewcommand{\arraystretch}{1.3} 
			\begin{tabular}{ccc}
				\toprule
				\multirow{3}[6]{*}{Method} & \multicolumn{2}{c}{Protocol-1} \\
				\cmidrule{2-3}      & Seen  & Unseen \\
				\cmidrule{2-3}      & ACC   & ACC \\
				\midrule
				\multicolumn{1}{l}{SegFace w/ConNet } &    96.40   & 47.09  \\
				\multicolumn{1}{l}{SegFace w/ResNet  } &    97.80   & 59.35 \\
				SegFace w/Swin-B  & 96.92 & 48.90 \\
				BiSeNet & \textbf{100 }  &\textbf{ 66.43} \\
				\bottomrule
			\end{tabular}%
		} 

	\end{minipage}
\end{table}

\begin{table}[t!]
	\centering
		\caption{Model ablation. ACC scores (\%) on Protocol-1. \label{tabab}}
	\setlength{\tabcolsep}{3.0mm}{
		\scriptsize
			\renewcommand{\arraystretch}{1.3} 
		\begin{tabular}{ccccccc}
			\toprule
			\multicolumn{5}{c}{\multirow{2}[2]{*}{Model}} & \multirow{2}[2]{*}{Seen} & \multirow{2}[2]{*}{Unseen} \\
			\multicolumn{5}{c}{}                 &       &  \\
			\midrule
			AE    & NE    & EE    & PE    & LE    & ACC   & ACC \\
			\midrule
			\multicolumn{1}{c}{\checkmark}    &       &       &       &       & 99.58 &\multicolumn{1}{c}{34.46}  \\
			&       & \multicolumn{1}{c}{\checkmark}    &       &       & 99.83 & 29.71 \\
			& \multicolumn{1}{c}{\checkmark}    &       &       &       & 99.66 & 23.97 \\
			\multicolumn{1}{c}{\checkmark}     & \multicolumn{1}{c}{\checkmark}   &       &       &       & 99.72 & 36.80 \\
			\multicolumn{1}{c}{\checkmark}    &\multicolumn{1}{c}{\checkmark}    &       &       & \multicolumn{1}{c}{\checkmark}     & 100   & 38.70 \\
			\multicolumn{1}{c}{\checkmark}     & \multicolumn{1}{c}{\checkmark}     &       & \multicolumn{1}{c}{\checkmark}     & \multicolumn{1}{c}{\checkmark}     & 100   & 61.64 \\
			\multicolumn{1}{c}{\checkmark}     &       & \multicolumn{1}{c}{\checkmark}     &       &       & 99.99 & 34.62 \\
			\multicolumn{1}{c}{\checkmark}   & \multicolumn{1}{c}{\checkmark}     & \multicolumn{1}{c}{\checkmark}     &       &       & 99.80 & 43.67 \\
			\multicolumn{1}{c}{\checkmark}     &\multicolumn{1}{c}{\checkmark}     & \multicolumn{1}{c}{\checkmark}    &  \multicolumn{1}{c}{\checkmark}       &    & 99.92 & 59.03 \\
			\multicolumn{1}{c}{\checkmark}&\multicolumn{1}{c}{\checkmark}    & \multicolumn{1}{c}{\checkmark}  &       & \multicolumn{1}{c}{\checkmark}   &100 & 61.25 \\
			\multicolumn{1}{c}{\checkmark}    & \multicolumn{1}{c}{\checkmark}   & \multicolumn{1}{c}{\checkmark}    & \multicolumn{1}{c}{\checkmark}  &\multicolumn{1}{c}{\checkmark }   &100   &66.43 \\
			\bottomrule
		\end{tabular}%
	}

	\centering
\end{table}	

\begin{figure}[t!]
	\centering
	\includegraphics[width=\linewidth]{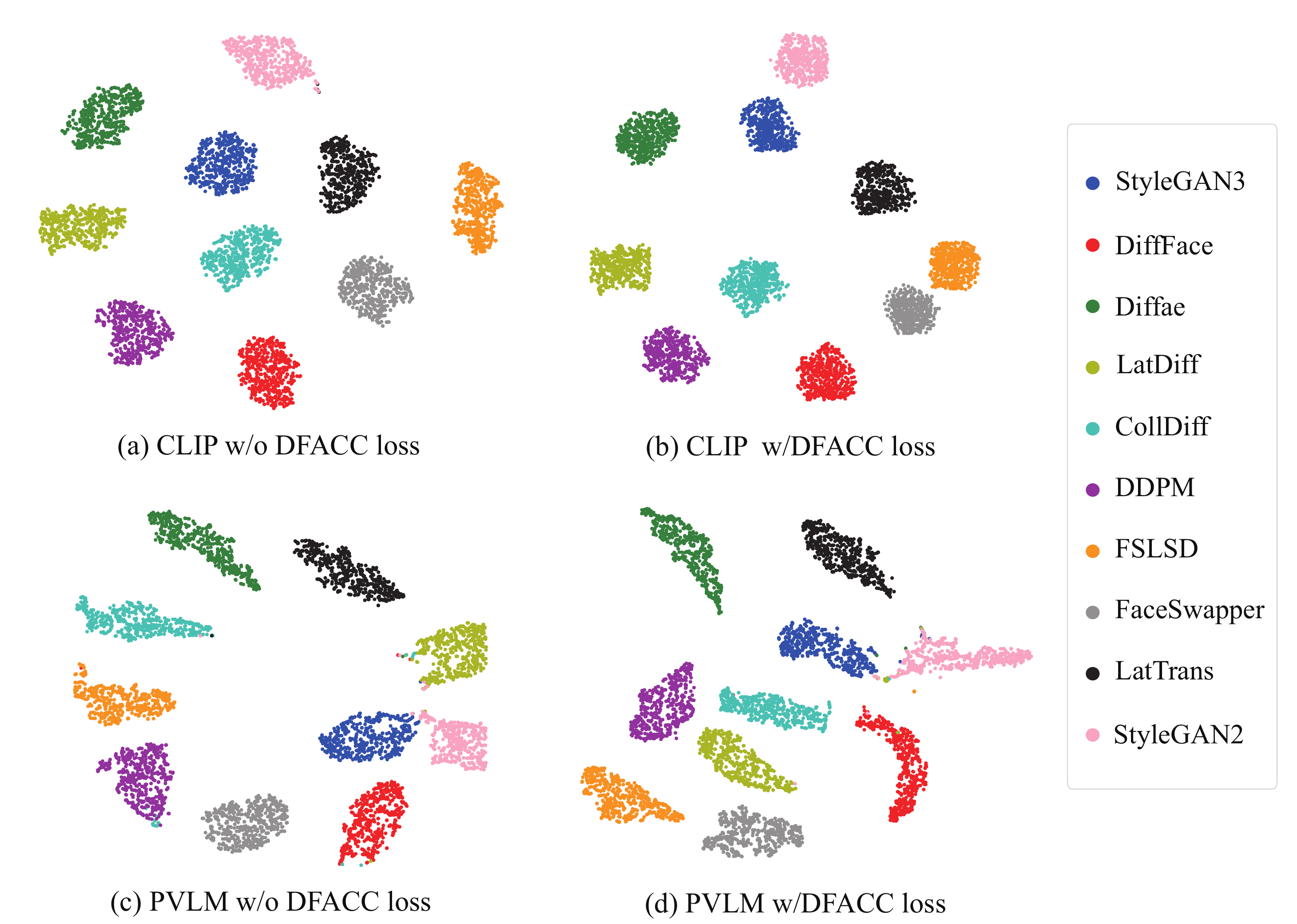} 
	\caption{The t-SNE visualization of various models (w/o or w/ DFACC loss).  }\label{figtsne}
\end{figure}

{\bfseries\setlength\parindent{0em} Impacts of the DFACC loss.} Models are trained with the VCC loss and the DFACC loss, respectively. In Table~\ref{abccloss}, the ACC of both CLIP and PVLM supervised by the VCC loss are 79.04\% and 96.90\% on protocol-1, respectively. By contrast, they achieve 89.07\% ACC and 100\% ACC under the constraints of our DFACC criteria, accordingly, which verifies the necessity of conducting intra-generator compactness and inter-generator separability based on the correlation among generators. We visualize features extracted from samples produced by various generators using t-SNE \cite{t-SNE}. We randomly select 500 samples for each seen generator from the testing set of protocol-1. In Fig.~\ref{figtsne}, samples of each generator are clustered in respective domains. With the addition of the DFACC loss, samples within the generator are more compact, with those of highly correlated generators aggregated and those of weakly related ones separated. This shows that our DFACC loss realizes flexible intra-generator aggregation and inter-generator separation.

{\bfseries\setlength\parindent{0em} Influences of the face parser.}
We delve into the impact of various face parsers, such as SegFace \cite{segface} and BiSeNet \cite{faceparser}. We test frameworks on protocol-1. In Table~\ref{tabfp}, the ACC reaches the maximum as the BiSeNet is involved. We argue that face parsing images extracted by BiSeNet are more discriminative than those encoded by other face parsers, which could boost DFA. We visualize parsing images generated by various face parsers, such as BiSeNet and SegFace, for fake faces produced by various generators. In Fig.~\ref{fig122}, compared to SegFace, the BiSeNet face parser does not generalize well to unseen diffusion faces. There are more evident distinctions among various generator parsing images produced by BiSeNet, visually. We attribute this to BiSeNet’s design as a lightweight face parsing network, optimized for real-world data. Unlike more complex models, such as GAN-based architectures or vision transformers like SegFace, BiSeNet may lack the capacity and flexibility to handle the unique characteristics of generative models and may also struggle with the high-frequency details and artifacts typical of diffusion models.

 We further calculate the FID score of the various generator parsing images created by different face parsers. To ensure a fair and balanced evaluation, we maintain an equal distribution of genuine and forged face parsing images, at a 1:1 ratio. We use 10,000 real images and 10,000 fake images to calculate the FID score for each generator. Lower FID scores denote better image quality. Results across various generators are shown in Fig.~\ref{fig122}. The FID score of diffusion parsing images from BiSeNet is higher than that of those from SegFace, showing that BiSeNet struggles to generalize to unseen diffusion faces. Despite the performance degradation of BiSeNet, in Table~\ref{tabfp}, the ACC score of the model with BiSeNet is higher than that of the model with SegFace. We argue that there are evident distinctions among various generator parsing distributions produced by BiSeNet, which can facilitate DFA. 

\begin{figure*}[t!]
	\centering
	\includegraphics[width=\linewidth]{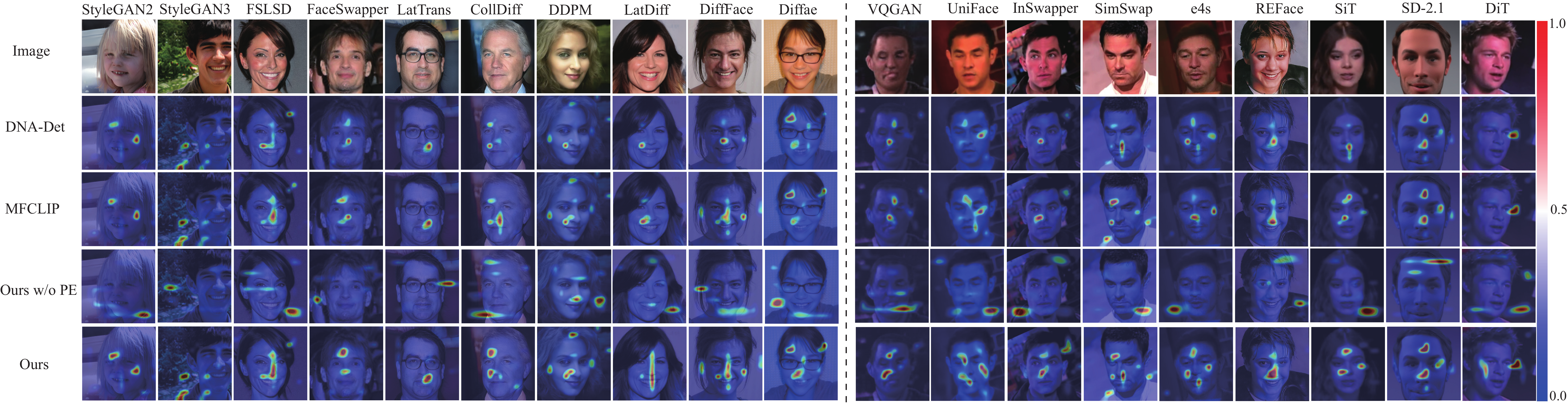} 
	\caption{The heatmap visualization of various models on the sample created by the seen (Left) or unseen (Right) generators. The hotter (red color) a position is, the more forgery traces are captured by the models.}\label{figheat}
\end{figure*}

{\bfseries\setlength\parindent{0em}Effects of the parsing encoder.} To intuitively demonstrate the effectiveness of PE, we visualize the heatmap of our method with or without PE, respectively, using Grad-CAM \cite{gradcam}. In Fig.~\ref{figheat}, we show the heatmap of various generator samples for four models. Each row displays a fake face generated by various seen and unseen generators. The second to fifth rows display heatmaps for four DFA models: DNA-Det, MFCLIP, PVLM without PE, and our PVLM method. PVLM without PE tends to overfit to small local areas or concentrate on semantic content noises, such as background outside the face area. With the introduction of PE, our method focuses on capturing long-range salient forgery features in the face, such as the nose and mouth, on both the seen and unseen generators. This verifies that our model is capable of suppressing semantic content noises and exploring common generator-specific forgery traces.

\begin{figure}[t!]
	\centering
	\includegraphics[width=\linewidth]{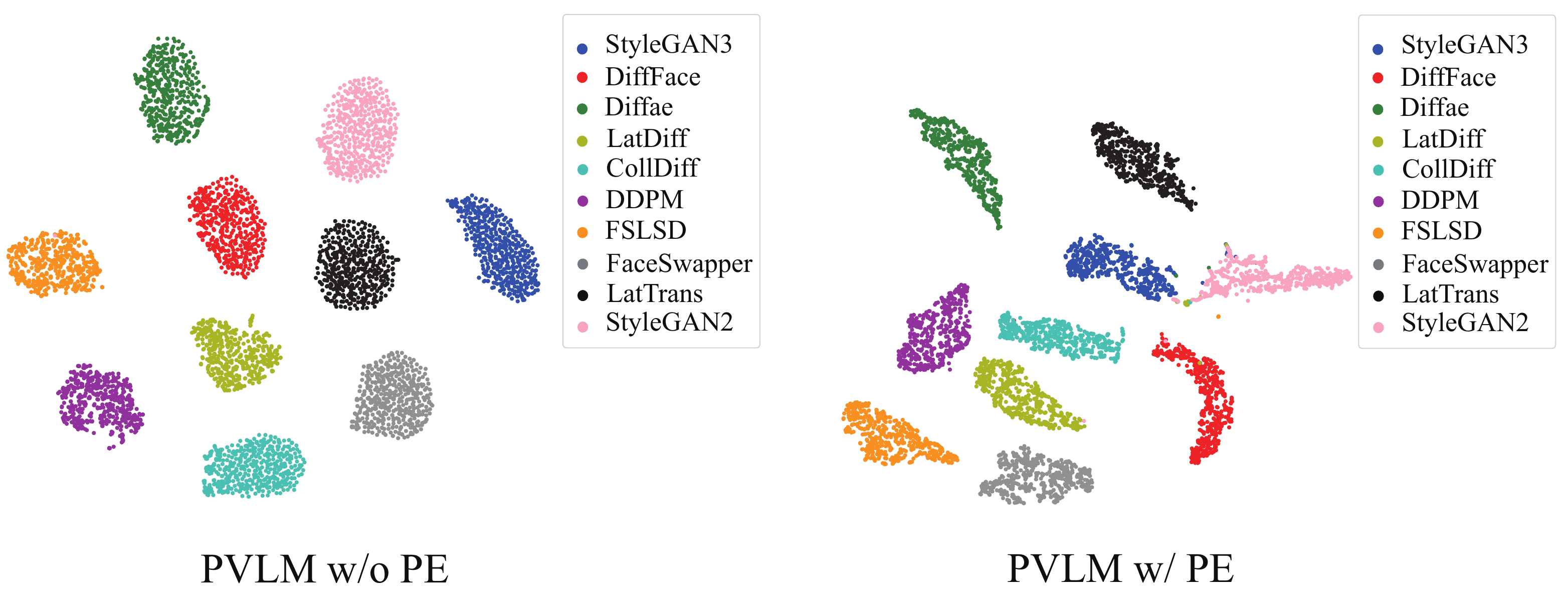} 
	\caption{The t-SNE feature visualization of our models (w/o or w/ PE) on the sample created by different generators.}\label{figpet}
\end{figure}

To further verify the generalization advantages of PE, we visualize the features extracted by our model (with or without PE) from samples generated by various generators using t-SNE \cite{tsne}. We pick 500 samples for each generator. In Fig.~\ref{figpet}, features of various generators encoded by the PVLM without PE are clustered in different domains. With the introduction of PE, the sample features within the generator become more compact, and the samples across generators from the same manipulation start to gather together, like Diffae and Lattrans from the AM manipulation, as well as StyleGAN2 and StyleGAN3 from the EFS forgery. In Table~\ref{tabab}, the ACC of our model without PE is 61.25\% on unseen generators, and the improvement from adding PE (+5.3\%) is evident. This proves that facial parsing-aware guidance could facilitate models to extract more general attribution features.

{\bfseries\setlength\parindent{0em} Impacts of different losses.} To delve into the contribution of loss functions, we conduct ablations on different losses. The ablation result of various losses is displayed in Table~\ref{tablosss}. When PVLM is supervised with purely DFA loss, the ACC is 35.76\% on unseen generators, but an increase of about 4.4\% in ACC could be achieved via adding the DFACC criterion. We argue that it could boost the generalizable deepfake attribution representations via flexible contrastive learning across various generators. At the same time, owing to the addition of the CMC loss, PVLM is improved by 20.05\% ACC on unseen generators, demonstrating the effectiveness of language guidance. The introduction of the DCPC loss further increases the performance (+5.5\%), which shows that the global face parsing features could be valuable prior information for both deepfake detection and attribution. Moreover, due to the integration of the KL loss, our PVLM model achieves an enhancement of 0.7\% ACC, which indicates that it is valuable to conduct feature alignment between visual and textual modalities. The incorporation of the five criteria yields the best performance among these losses, indicating that the proposed loss realizes promising results.

{\bfseries\setlength\parindent{0em} Effects of the margin $m$ in DFACC loss.} In the DFACC loss, we measure the correlation between generators based on the margin value. We investigate the influence of different margin values on model performance. As shown in Table~\ref{mar}, the ACC score tends to improve with an increase of the margin value, indicating that a larger margin enhances the model's ability to distinguish between generators, effectively. Notably, the performance reaches the peak when the margin value is set to 0.7. This suggests that a margin of 0.7 can serve as an optimal value for evaluating the correlation among generators.

{\bfseries\setlength\parindent{0em} Influences of the language encoder.}
To delve into the impact of LE, we conduct ablation experiments using protocol-1. The ablation results are shown in Table~\ref{table}. The gains from introducing the LE module (+7.4\%) are obvious. We argue that LE offers fine-grained and accurate language guidance via vision-language matching, which could enhance the DFA performance of models.

{\bfseries\setlength\parindent{0em} Impacts of the MMI module.}
To study the effect of MMI, we conduct ablation experiments using protocol-1. The ablation results are shown in Table~\ref{tabloss}. The gains from adding the MMI module (+3.26\%) are evident. We argue that MMI facilitates diverse and comprehensive multi-view interactions, promoting the model to extract rich attribution features.

{\bfseries\setlength\parindent{0em} Effects of the fine-grained text prompts.} To study the impact of fine-grained text prompts on model performance, we introduce learnable prompts with insufficient accuracy, to train PVLM and CLIP, respectively. In Table~\ref{ab4}, the performance of both CLIP and PVLM declines significantly (-25.52\% and -17.42\%), due to the addition of learnable text prompts, accordingly. We argue that elaborate and accurate semantic guidance plays a vital role in fine-grained DFA tasks.

\begin{table}[t]
	\begin{minipage}{0.46\linewidth}
			\caption{Effects of the MMI module. \label{tabloss}}
		\centering
		\setlength{\tabcolsep}{1.4mm}{
			\scriptsize
			\renewcommand{\arraystretch}{1.3} 
			\begin{tabular}{rcc}
				\toprule
				\multicolumn{1}{c}{\multirow{3}[6]{*}{Method}} & \multicolumn{2}{c}{Protocol-1} \\
				\cmidrule{2-3}      & Seen  & Unseen \\
				\cmidrule{2-3}      & ACC   & ACC \\
				\midrule
				\multicolumn{1}{l}{	Ours w/o MMI} & 99.60 &  63.17\\
				\multicolumn{1}{l}{Ours w/ MMI} &    \textbf{100}&    \textbf{66.43} \\
				\bottomrule
			\end{tabular}%
		
		}
	
		\vspace{-1em}
	\end{minipage}
	\hfill
	\begin{minipage}{0.47\linewidth}  
		\centering		
			\caption{Effects of the language encoder.\label{table}}
		\setlength{\tabcolsep}{0.9mm}{
			\scriptsize
				\renewcommand{\arraystretch}{1.3} 
			\begin{tabular}{ccc}
				\toprule
				\multirow{3}[6]{*}{Method} & \multicolumn{2}{c}{Protocol-1} \\
				\cmidrule{2-3}      & Seen  & Unseen \\
				\cmidrule{2-3}      & ACC   & ACC \\
				\midrule
				\multicolumn{1}{l}{Ours w/o LE   } &    99.92   &  59.03 \\
				\multicolumn{1}{l}{Ours w/ LE } &    \textbf{100}   & \textbf{66.43} \\
				\bottomrule
			\end{tabular}%
		} 
	
	\end{minipage}
\end{table}

\begin{table}[t]
		\caption{Effects of various losses. \label{tablosss}}
	\centering
	\setlength{\tabcolsep}{1.4mm}{
		\scriptsize
			\renewcommand{\arraystretch}{1.3} 
		\begin{tabular}{rcc}
			\toprule
			\multicolumn{1}{c}{\multirow{3}[6]{*}{Method}} & \multicolumn{2}{c}{Protocol-1} \\
			\cmidrule{2-3}      & Seen  & Unseen \\
			\cmidrule{2-3}      & ACC   & ACC \\
			\midrule
			\multicolumn{1}{l}{	$\mathcal{L}_\text{dfa} $ } & 99.32 &  35.76\\
			\multicolumn{1}{l}{$\mathcal{L}_\text{dfa} +\mathcal{L}_\text{dfacc}$} & 99.74 & 40.16 \\
			\multicolumn{1}{l}{$\mathcal{L}_\text{dfa} +\mathcal{L}_\text{dfacc} +\mathcal{L}_\text{cmc}$}&     99.98  &60.21  \\
			\multicolumn{1}{l}{$\mathcal{L}_\text{dfa} +\mathcal{L}_\text{dfacc} +\mathcal{L}_\text{cmc}+ \mathcal{L}_\text{dcpc}$}&     100  &  65.73\\
			$\mathcal{L}_\text{dfa} +\mathcal{L}_\text{dfacc} +\mathcal{L}_\text{cmc}+ \mathcal{L}_\text{dcpc}+\mathcal{L}_\text{kl}$ &      \textbf{100} &    \textbf{66.43}\\
			\bottomrule
		\end{tabular}%
	}

\end{table}

\begin{table}[t]
	\caption{Effects of various priors. \label{pe}}
	\centering
	\setlength{\tabcolsep}{2.5mm}{
		\scriptsize
		\renewcommand{\arraystretch}{1.3} 
		\begin{tabular}{rcc}
			\toprule
			\multicolumn{1}{c}{\multirow{3}[6]{*}{Method}} & \multicolumn{2}{c}{Protocol-1} \\
			\cmidrule{2-3}      & Seen  & Unseen \\
			\cmidrule{2-3}      & ACC   & ACC \\
			\midrule
					\multicolumn{1}{l}{PE w/ edge } & 99.83 & 29.71 \\
			\multicolumn{1}{l}{PE w/ appearance}&     99.58  &34.46  \\
			\multicolumn{1}{l}{	PE w/ parsing } & 99.42 &  37.06\\
			\bottomrule
		\end{tabular}%
	}
	
\end{table}

\begin{table}[t]
	\centering
	\caption{Effects of text prompts. LP is the learnable prompt.}	\label{ab4}
	\setlength{\tabcolsep}{4.3mm}{
		\scriptsize
		\renewcommand{\arraystretch}{1.3} 
		\begin{tabular}{lcc}
			\toprule
			\multicolumn{1}{c}{\multirow{3}[6]{*}{Method}} & \multicolumn{2}{c}{Protocol-1} \\
			\cmidrule{2-3}      & Seen  & Unseen \\
			\cmidrule{2-3}      & ACC   & ACC \\
			\midrule
			CLIP w/LP & 53.52&-  \\
			CLIP w/o LP & \textbf{79.04} &-  \\
			
			\midrule
			Ours w/ LP & 82.58   &45.79 \\
			Ours w/o LP & \textbf{100}  & \textbf{66.43}  \\
			\bottomrule
		\end{tabular}%
	}
\vspace{-1em}
\end{table}

{\bfseries\setlength\parindent{0em}Influences of the factor $\lambda$ in DFACC loss.}  In the DFACC loss, we define the factor $\lambda$ to balance the loss term between $\mathcal{L}_\mathrm{intra}$ and $\mathcal{L}_\mathrm{inter}$. We study the effect of various factors on model performance. As Table~\ref{fac} displays, the ACC score rises with the increase of factors, which indicates the effectiveness of flexible metric learning. Specifically, the performance reaches its peak when the weight factor $\lambda$ is set to 0.5, which means an optimal balance between the two loss terms. We argue that the optimal setting of $\lambda$ allows the model to learn the most discriminative features, leading to better generalization and more accurate predictions.

{\bfseries\setlength\parindent{0em}Effects of various priors.}To verify the effectiveness of face parsing priors and PE, various image priors such as appearance images, face parsing images, or edge images, are fed into the PE and then passed to the MLP head to generate attribution logits, which are supervised only by the DFA loss. We train models using generators in protocol-1 and test them on seen and unseen generators, respectively. In Table~\ref{pe},  the ACC of PE with the parsing prior is 99.42\% and 37.06\% on seen and unseen generators, respectively, which shows that PE may preserve the generator-specific features. Besides, the ACC of PE with the parsing prior is nearly 2.6\% higher than that of PE with the appearance prior on unseen generators. This shows that parsing priors are more advantageous than other priors in improving model generalization. We argue that they exhibit more pronounced distributional differences, which boosts PE to extract the common generator-specific forgery features.
\begin{table}[t]	
	\centering
		\caption{Influence of margin values in DFACC loss.}
	\setlength{\tabcolsep}{2.80mm}{
		\scriptsize
			\renewcommand{\arraystretch}{1.3} 
		\begin{tabular}{clcc}
			\toprule
			\multirow{3}[6]{*}{Method} & \multirow{3}[6]{*}{} & \multicolumn{2}{c}{Protocol-1} \\
			\cmidrule{3-4}      &       & Seen  & Unseen \\
			\cmidrule{3-4}      &       & ACC   & ACC \\
			\midrule
			\multirow{3}[2]{*}{CLIP} & m=0.5 & 88.85 & 38.21 \\
			& m=0.7 & 89.07 &40.97 \\
			& m=0.9 & 79.01 & 34.06 \\
			\midrule
			\multirow{3}[2]{*}{PVLM (Ours)} & m=0.5 & 98.02 & 49.36 \\
			& m=0.7 & \textbf{100}   & \textbf{66.43} \\
			& m=0.9 & 99.40 & 54.98 \\
			\bottomrule
		\end{tabular}%
	}
\vspace{-1em}
	\label{mar} 
\end{table}

\begin{table}[t]
	
	\centering
		\caption{Influence of the factor in DFACC loss.}
	\setlength{\tabcolsep}{2.80mm}{
		\scriptsize
			\renewcommand{\arraystretch}{1.3} 
		\begin{tabular}{clcc}\toprule\multirow{3}[6]{*}{} & \multicolumn{1}{c}{\multirow{3}[6]{*}{Method}} & \multicolumn{2}{c}{Testing Set} \\\cmidrule{3-4}      &       & Seen  & Unseen \\\cmidrule{3-4}      &       & ACC   & ACC \\\midrule\multirow{4}[2]{*}{CLIP} &  $\lambda$=0.0   & 79.46 &  34.89\\      &  $\lambda$=0.3 & 82.12 & 35.60 \\      &  $\lambda$=0.5 & 88.85 &38.21  \\      &  $\lambda$=0.7 & 87.86 & 37.65\\\midrule\multirow{4}[2]{*}{PVLM  (Ours)} &  $\lambda$=0.0   & 94.37 & 62.24 \\      &  $\lambda$=0.3 & 97.46 & 64.02 \\      &  $\lambda$=0.5 &  \textbf{100}   &  \textbf{66.43} \\      &  $\lambda$=0.7 & 98.03 & 64.28 \\\bottomrule\end{tabular}%
	}
\vspace{-1em}
	\label{fac} 
\end{table}

\begin{figure*}[t]
	\centering
	\includegraphics[width=0.9\linewidth]{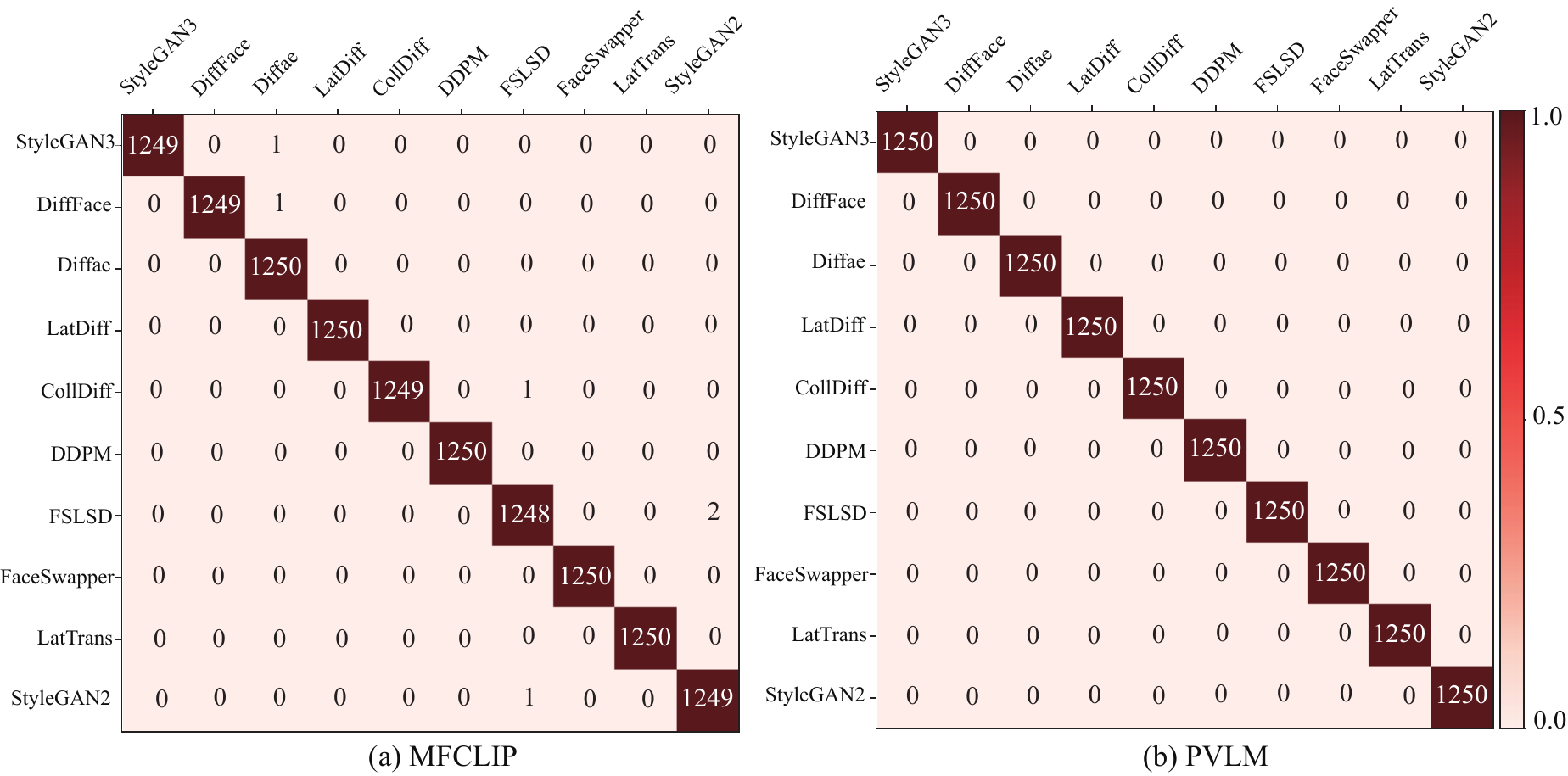} 
	\caption{Confusion matrix visualization of various methods. The darker red means more frequent or stronger predictions, and the lighter red denotes less frequent or weaker predictions. The entries of the confusion matrix represent the number of samples.}\label{fig7}
\end{figure*}

\section{Visualization }
{\bfseries\setlength\parindent{0em} Visualization of confusion matrix.} We visualize the confusion matrix to observe the predictions of different models. As shown in Fig.~\ref{fig7}, MFCLIP achieves lower prediction results compared to PVLM. In detail, some samples may be misclassified, particularly in the following cases: when the data source is identical, such as StyleGAN2 and FSLSD, or when they can be categorized into the same classes, such as the diffusion-based model (e.g., DiffFace and Diffae). By contrast, our PVLM method is capable of decreasing confusion between similar generators while achieving more accurate predictions, i.e., 100\% ACC on all generators.

{\bfseries\setlength\parindent{0em} Visualization of heatmaps.}To enhance the interpretability of the proposed method, we visualize the heatmap yielded by various settings using the Grad-CAM \cite{Grad-CAM}. In Fig.~\ref{figheat}, we show the heatmap of samples generated by various seen and unseen generators for four models, respectively. Each column shows a fake face synthesized by different generators. The second to fourth rows display heatmaps for four DFA models: DNA-Det, MFCLIP, PVLM without PE, and our PVLM method. It is observed that PVLM mines more comprehensive forgery traces than both MFCLIP and DNA-Det. As shown in Fig.~\ref{figheat}, attributors tend to identify forgery regions such as eyes, nose, and mouth. It is noted that DFA could be regarded as a fine-grained forgery detection task, since manipulated artifacts created by generators are prone to being subtle and faint.
\vspace{-1.1em}
\section{Deepfake Detection}
To study the significance of DFA, we evaluate the performance of the deepfake detectors and the deepfake attributors model using Protocol-2, respectively. The results of models on two tasks (i.e., deepfake detection and deepfake attribution) are displayed in Table~\ref{tabdfd}. It is noticed that the classification tends to become easier, with the number of categories increasing from 2 (deepfake detection) to 11 (deepfake attribution). We argue that there is a growing distinction between pristine and forged images, as the number of categories rises, which facilitates deepfake detection. For example, MFCLIP and DNA-Det obtain 99.74\% ACC and 96.18\% ACC on seen generators under the deepfake detection setting, respectively. By contrast, they achieve 99.91\% ACC (+0.17\%) and 99.65\% ACC (+2.47\%) on seen generators in DFA tasks, accordingly.
\begin{table}[t!]
		\centering
	\caption{ACC results (\%) of models for deepfake detection and deepfake attribution on protocol-2. \label{tabdfd}}
	\setlength{\tabcolsep}{2.3mm}{
		\scriptsize
		\renewcommand{\arraystretch}{1.3} 
		\begin{tabular}{crrrr}
			\toprule
			\multirow{3}[6]{*}{Method} & \multicolumn{2}{c}{Deepfake Detection} & \multicolumn{2}{c}{Deepfake Attribution} \\
			\cmidrule{2-5}      & \multicolumn{1}{c}{Seen} & \multicolumn{1}{c}{Unseen} & \multicolumn{1}{c}{Seen} & \multicolumn{1}{c}{Unseen} \\
			\cmidrule{2-5}      & \multicolumn{1}{c}{ACC} & \multicolumn{1}{c}{ACC} & \multicolumn{1}{c}{ACC} & \multicolumn{1}{c}{ACC} \\
			\midrule
			CLIP \cite{clip} &   \multicolumn{1}{c}{ 73.56 }  &   \multicolumn{1}{c}{63.08}    &   \multicolumn{1}{c}{ 82.72}   &\multicolumn{1}{c}{ 35.98} \\
			CAEL \cite{genface}  & \multicolumn{1}{c}{99.63} & \multicolumn{1}{c}{83.09} & \multicolumn{1}{c}{99.83} &  \multicolumn{1}{c}{30.86} \\
			MFCLIP \cite{MFCLIP} &   \multicolumn{1}{c}{99.74}    &    \multicolumn{1}{c}{86.49}   &   \multicolumn{1}{c}{99.91}    & \multicolumn{1}{c}{43.38} \\
			DE-FAKE \cite{DEFAKE} & \multicolumn{1}{c}{97.07} & \multicolumn{1}{c}{69.32} & \multicolumn{1}{c}{98.15}     & \multicolumn{1}{c}{9.30}\\
			DNA-Det \cite{DNADet} & \multicolumn{1}{c}{96.18} &  \multicolumn{1}{c}{ 67.02}    & \multicolumn{1}{c}{99.65} & \multicolumn{1}{c}{50.48} \\
			PVLM (Ours)&   \multicolumn{1}{c}{\textbf{ 99.90}}    &  \multicolumn{1}{c}{ \textbf{ 89.45} }   &  \multicolumn{1}{c}{ \textbf{100}  }   & \multicolumn{1}{c}{ \textbf{67.61} } \\
			\bottomrule
		\end{tabular}%
	}
	
\end{table}

\section{Conclusion}
In this paper, we conduct a novel fine-grained ZS-DFA benchmark. Besides, we propose an innovative parsing-aware vision-language model with dynamic contrastive learning (PVLM) method, which introduces face parsing modality to enhance the learning of visual forgery features across image, noise, and edge views, thus mining both general and discriminative attribution patterns, to achieve ZS-DFA. We demonstrate the importance of face parsing priors and the PE module for our ZS-DFA task from the theoretical perspective of signal decoupling. We prove that the parsing encoder improves the model's generalization ability to unseen generators by suppressing semantic noises and focusing on general generator-specific fingerprints. This theoretical framework provides clear design principles and new insights for researchers in improving current models and exploring new methods to alleviate bias. Although our method has achieved excellent ZS-DFA, the interaction among vision, parsing, and language modalities is limited. We intend to explore diverse and efficient vision-language-parsing communication mechanisms to improve the ZS-DFA performance of PVLM. We also plan to introduce abundant face data generated by various generators like flow-based and auto-regressive models, to facilitate ZS-DFA. We hope that our new benchmark and PVLM model could provide new insights for the research direction, such as deepfake attribution, deepfake detection, and vision-language model.

\bibliographystyle{IEEEtran}
\bibliography{BM2PVL}

\begin{IEEEbiography}[{\includegraphics[width=1in,clip,keepaspectratio]{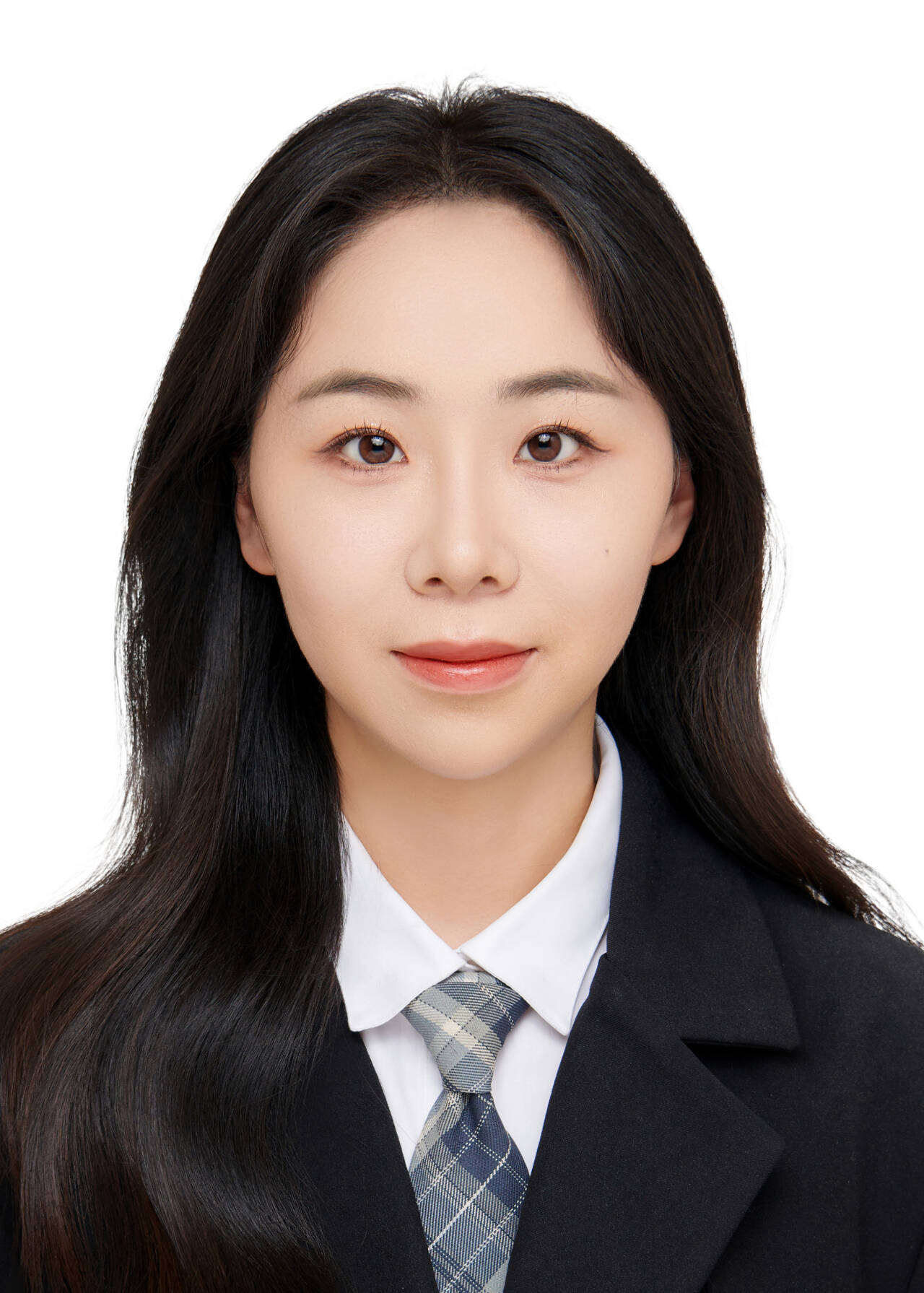}}]{Yaning Zhang} received the double bachelor’s degree in Internet of Things Engineering and English and the M.S. degree in Computer Applied Technology from Qilu University of Technology (Shandong Academy of Sciences), Jinan, China, in 2020 and 2023, respectively, where she is currently pursuing the Ph.D. degree. Her research interests include computer vision, artificial intelligence, multimedia forensics, and face forgery detection.
\end{IEEEbiography}

\begin{IEEEbiography}[{\includegraphics[width=1in,clip,keepaspectratio]{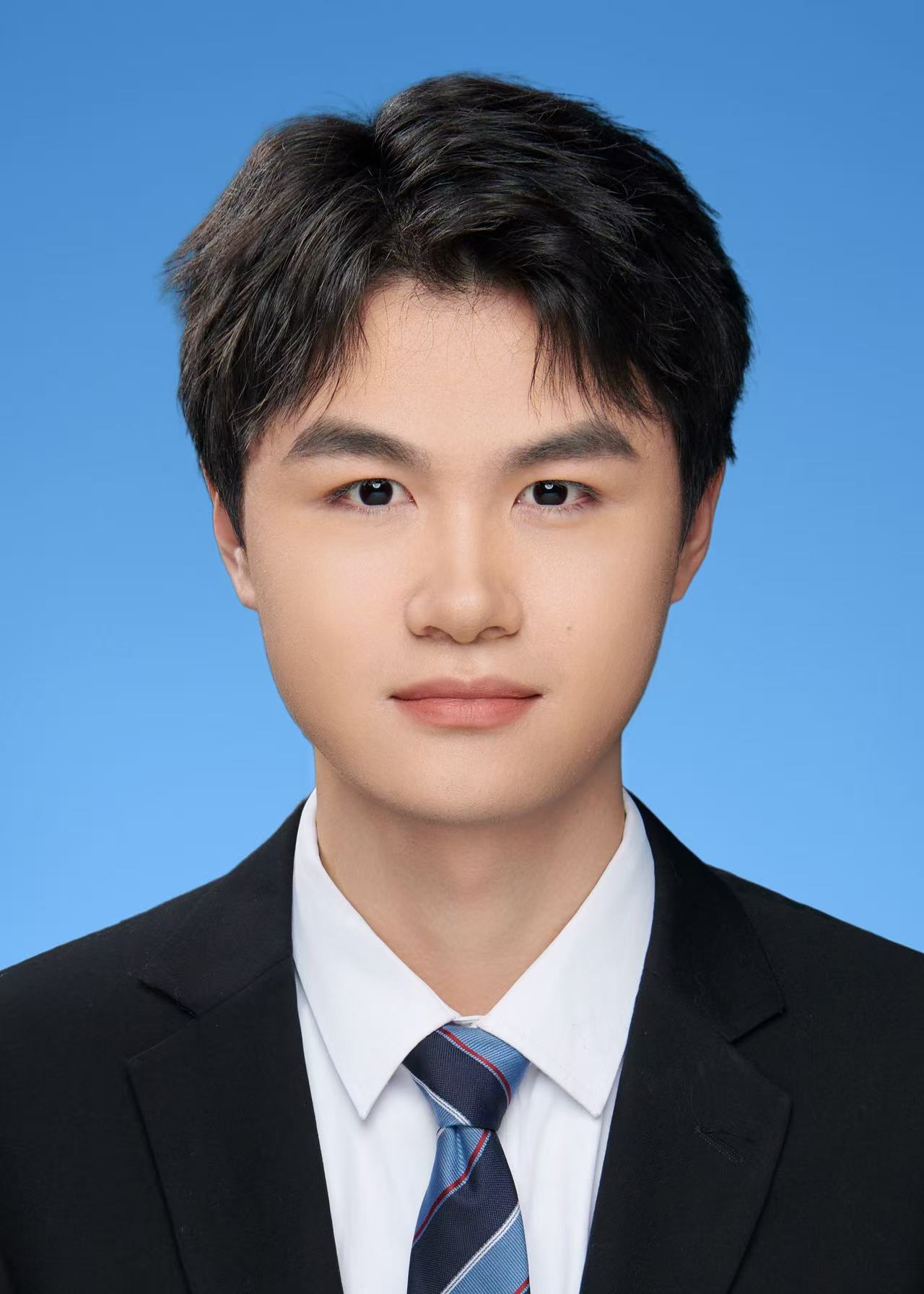}}]{Jiahe Zhang} is currently pursuing the B.E. degree at Shandong University of Science and Technology, Jinan, China. His major research interests include computer vision, artificial intelligence, and machine learning. 
\end{IEEEbiography}
\begin{IEEEbiography}[{\includegraphics[width=1in,clip,keepaspectratio]{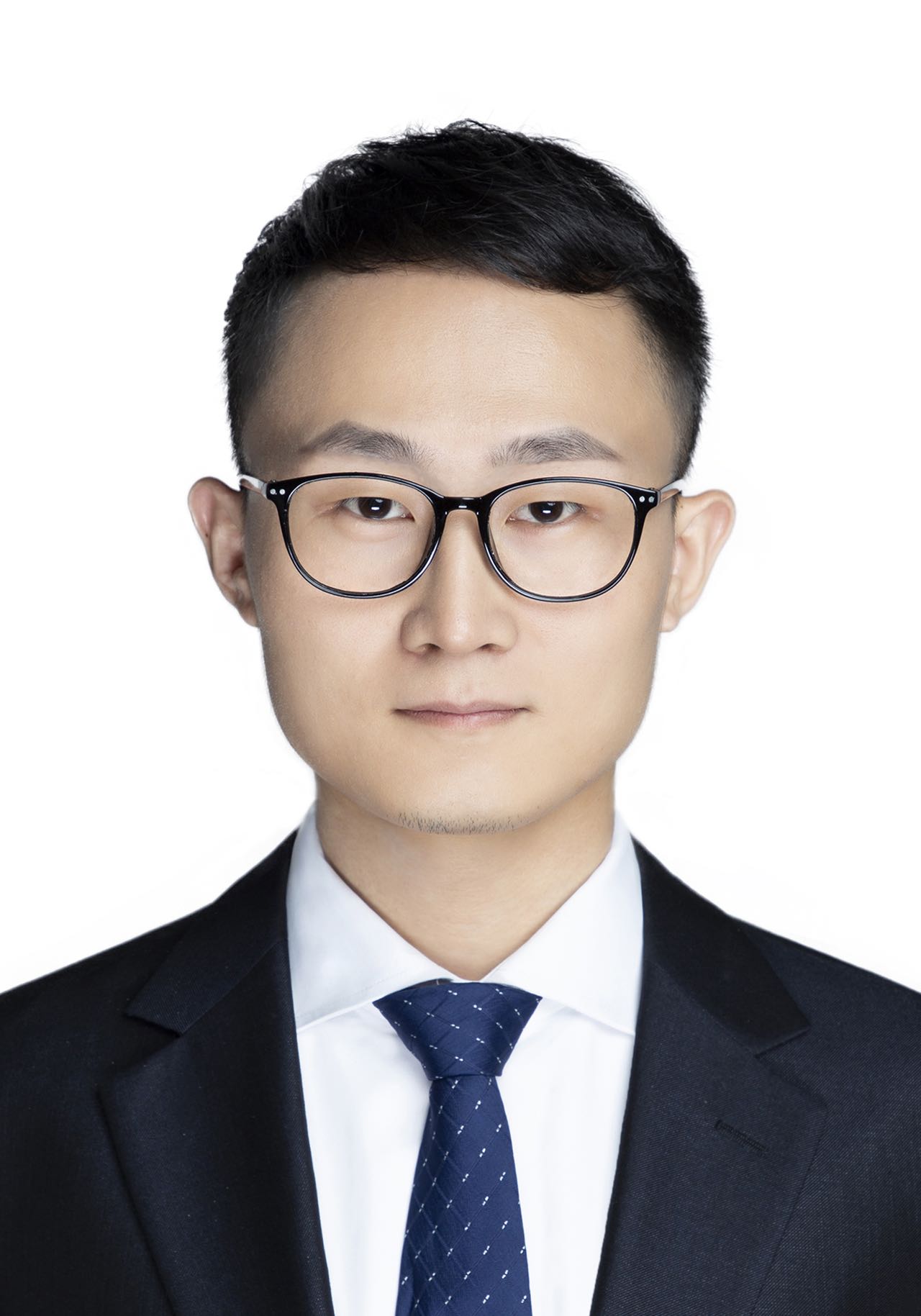}}]{Chunjie Ma} received the Ph.D. degree in Electronic Science and Technology from the Beijing University of Technology in 2023. He is currently a Research Assistant with the Shandong Artificial Intelligence Institute, Qilu University of Technology. His research interests include computer vision, X-ray image prohibited object detection, tamper detection, etc.
\end{IEEEbiography}

\begin{IEEEbiography}[{\includegraphics[width=1in,clip,keepaspectratio]{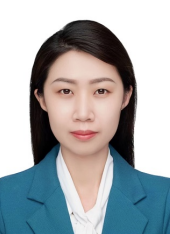}}]{Weili Guan} (Member, IEEE) received the master’s degree from the National University of Singapore, and the PhD degree from Monash University. She has about 6 years of working experience at the enterprise. She is currently a professor with the School of Electronics and Information Engineering, Harbin Institute of Technology (Shenzhen), China. Her research interests are multimedia computing and information retrieval. She has published more than 60 papers at the first-tier conferences and journals, like ACM MM, SIGIR, IEEE Transactions on Pattern Analysis and Machine Intelligence and IEEE Transactions on Image Processing.
\end{IEEEbiography} 

\begin{IEEEbiography}[{\includegraphics[width=1in,clip,keepaspectratio]{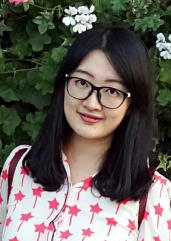}}]{Tian Gan} (Member, IEEE) received the B.Sc. degree from East China Normal University in 2010 and the Ph.D. degree from the National University of Singapore in 2015. She was a Research Scientist with the Institute for Infocomm Research (I2R), Agency for Science, Technology and Research (A*STAR). She is currently an Associate Professor with the School of Computer Science and Technology, Shandong University. Her research interests include social media marketing, video understanding, and multimedia computing.
\end{IEEEbiography} 

\begin{IEEEbiography}[{\includegraphics[width=1in,clip,keepaspectratio]{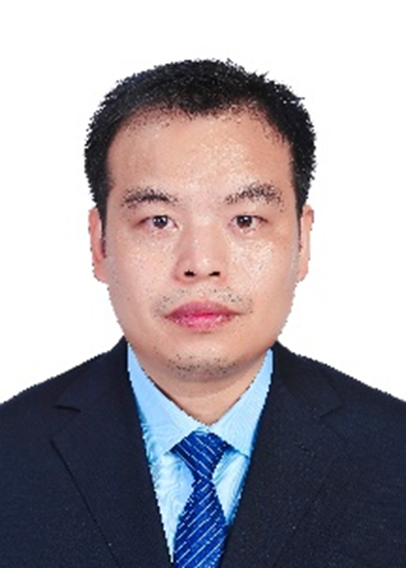}}]{Zan Gao}  (Senior Member, IEEE) received his Ph.D degree from Beijing University of Posts and Telecommunications in 2011. He is currently a full Professor with the Shandong Artificial Intelligence Institute, Qilu University of Technology (Shandong Academy of Sciences). From Sep. 2009 to Sep. 2010, he worded in the School of Computer Science, Carnegie Mellon University, USA. From July 2016 to Jan 2017, he worked in the School of Computing of National University of Singapore. His research interests include artificial intelligence, multimedia analysis and retrieval, and machine learning. He has authored over 100 scientific papers in international conferences and journals including TPAMI, TIP, TNNLS, TMM, TCYBE, CVPR, ACM MM, WWW, SIGIR and AAAI.
\end{IEEEbiography}

\end{document}